\PassOptionsToPackage{table}{xcolor}
\documentclass{youtu}
\usepackage[T1]{fontenc}
\usepackage{mathpazo}
\usepackage{natbib}
\setcitestyle{numbers,square}

\usepackage{amsmath,amsfonts,bm}

\def\eqref#1{equation~\ref{#1}}

\def\1{\bm{1}}

\DeclareMathAlphabet{\mathsfit}{\encodingdefault}{\sfdefault}{m}{sl}
\SetMathAlphabet{\mathsfit}{bold}{\encodingdefault}{\sfdefault}{bx}{n}

\usepackage{graphicx}
\usepackage{tikz}
\usetikzlibrary{calc,arrows.meta,decorations.pathreplacing,positioning}
\usepackage{subfigure}
\usepackage{wrapfig}
\usepackage{needspace}
\usepackage{placeins} %
\usepackage{tcolorbox}
\usepackage{scalerel}

\usepackage{amsmath}
\usepackage{amstext}
\usepackage{amsfonts}
\usepackage{amssymb}
\usepackage{bm}
\usepackage{bbm}
\usepackage[mathscr]{euscript}

\usepackage{multirow}
\usepackage{booktabs}
\usepackage{array}
\usepackage{tabularx}
\usepackage{caption}
\usepackage[table]{xcolor}
\usepackage{tablefootnote}
\usepackage{adjustbox}
\usepackage{fix-cm}
\usepackage{nicematrix}
\usepackage{algorithm}
\usepackage{algpseudocode}

\usepackage{enumitem}
\usepackage{soul}
\usepackage{pifont}
\usepackage{listings}
\usepackage{xspace}
\usepackage[normalem]{ulem}

\usepackage{url}
\usepackage{hyperref}

\newcommand{\method}{\textsc{Attuner}\xspace}
\newcommand{\pic}{PIC\xspace}

\algnewcommand{\LineComment}[1]{\Statex ~~~~~~\textsc{//}~\textit{#1}}

\setenumerate[1]{itemsep=0pt,partopsep=0pt,parsep=\parskip,topsep=5pt}
\setitemize[1]{itemsep=0pt,partopsep=0pt,parsep=\parskip,topsep=5pt}
\setdescription{itemsep=0pt,partopsep=0pt,parsep=\parskip,topsep=5pt}

\definecolor{c0}{cmyk}{1,0.3968,0,0.2588}
\definecolor{LightCyan}{rgb}{0.88,1,1}
\definecolor{TableHeader}{HTML}{EEF2F5}
\definecolor{ReuseColumn}{HTML}{EDF4FB}
\definecolor{FullGray}{HTML}{4B5563}
\definecolor{PICBlue}{HTML}{4C78A8}
\definecolor{GapOrange}{HTML}{E68632}
\definecolor{UtilTeal}{HTML}{2A9D8F}
\definecolor{OraclePurple}{HTML}{7A5195}
\definecolor{MuSiQueTint}{HTML}{FFF3E6}
\definecolor{BigCodeTint}{HTML}{EDF4FB}
\definecolor{SkillTint}{HTML}{EDF4FB}
\definecolor{DocumentTint}{HTML}{FFF3E6}
\definecolor{MemoryTint}{HTML}{EAF6F1}
\definecolor{RepairGray}{HTML}{D9DEE3}

\definecolor{magenta}{rgb}{1.0,0.0,0.56}

\definecolor{uclablue}{rgb}{0.15,0.45,0.68}
\hypersetup{
    breaklinks,
    citecolor=uclablue,
    colorlinks=true,
    linkcolor=uclablue,
    urlcolor=uclablue
}

\title{
\method: Recomputation-Free Agentic Context KV Reuse
}

\author{\small Xinghao Chen$^{1,2,3\dagger}$, Junnan Dong$^{3\dagger}$, Cai Ke$^{3}$, Chak Tou Leong$^{2}$, Haocheng Sun$^{3}$,\\
Keyu Chen$^{3}$, Siyu An$^{3}$, Ruizhi Qiao$^{3}$, Xing Sun$^{3}$, Wenjie Li$^{2}$, Xiaoyu Shen$^{1\ddagger}$}
\affiliation{$^{1}$EIT-NLP Lab, Eastern Institute of Technology, Ningbo\\
$^{2}$The Hong Kong Polytechnic University\qquad $^{3}$Tencent Youtu Lab}
\renewcommand{\contributionlist}{\begin{center}\small
$\dagger$ Equal contribution. \quad $\ddagger$ Corresponding author.\\
\texttt{xing-hao.chen@connect.polyu.hk}\quad\texttt{xyshen@eitech.edu.cn}
\end{center}}

\youtufinalcopy

\begin{document}

\abstract{

Large language model (LLM) agents repeatedly load reusable content, such as skills, documents, and memory entries, into the current context. Re-encoding this content for every request wastes computation. Position-independent caching (PIC) alleviates this by encoding each artifact independently and reusing its key-value (KV) states at arbitrary positions, but it incurs a quality loss relative to full-context prefill. Existing methods repair this loss by restoring global position IDs or recomputing selected tokens. In this work, we isolate the source of the loss, finding that the positional mismatch has minor effect, and independently cached artifacts retain faithful representations: reading a provided artifact stays largely accurate, and performance degrades only when the model must select among multiple artifacts.  Moreover, replacing PIC's attention scores with full-prefill scores recovers performance with the cached KV unchanged, localizing the failure to the attention rather than KV recomputation. Motivated by this, we propose \textsc{Attuner}, a query-side adaptation method that learns to read a frozen artifact cache.  \textsc{Attuner} inserts low-rank adapters into the query projections and is trained by distilling full-prefill distribution into the student. It trains fewer than 0.05\% of the model parameters and, at inference, requires neither cache recomputation nor a full-context reference. On Qwen3-4B and Qwen3-8B across seven benchmarks covering skills, documents, memory, and code, \textsc{Attuner} substantially outperforms prior PIC baselines in both in-domain and out-of-domain settings, matches full-context prefill quality while providing up to $3.73\times$ speedup.

}
\maketitle

\section{Introduction}
Large language model (LLM) agents repeatedly interact with external environments~\citep{liang2026unitoolcallunifyingtooluserepresentation}, assembling each request's context from reusable content such as skill descriptions, documents~\citep{dong2024knowgpt,dong2026youtu}, and memory entries~\citep{mei2025contextengineering,zhou2026agentskills,luo2026agentmemory,yao2023react,dong2026deep,ke2026disentanglinglongtermmemorylatent}.
We refer to these reusable spans as \emph{artifacts}. Such artifacts are stored persistently and loaded on demand at runtime~\citep{su2026skill}. Re-encoding them for every request wastes computation and raises time to first token (TTFT). Reusing key-value (KV) states avoids this cost~\citep{li2025kvcachesurvey}, but conventional prefix caching assumes reusable content occupies the same prefix position across requests~\citep{sglang}. Agentic workloads rarely satisfy this assumption: The artifacts to load, their order, and their positions are decided dynamically at runtime by retrieval, routing and planning.

Position-independent caching (PIC) addresses this limitation by encoding each artifact independently and loading cached blocks at request time~\citep{hu2024epic}. However, directly reusing independently constructed caches can incur a performance loss relative to full prefill~\citep{kvlink}. Two mismatches arise in PIC. The first is positional: artifacts are encoded before their position is known, so their position indices do not match those used in full prefill. The second is contextual: independently encoded artifacts attend do not attend to preceding request-specific artifacts, so their KV states lack the cross-artifact conditioning present in full prefill~\citep{yao2024cacheblend}. Existing methods mitigate these mismatches at runtime, either by restoring global positional embeddings~\citep{cacheslide,prophetkv} or by recomputing selected tokens to approximate the original full-context KV states~\citep{encoderpic,kvlink,sempic}.

\begin{figure}[t]
    \centering
    \includegraphics[width=\linewidth]{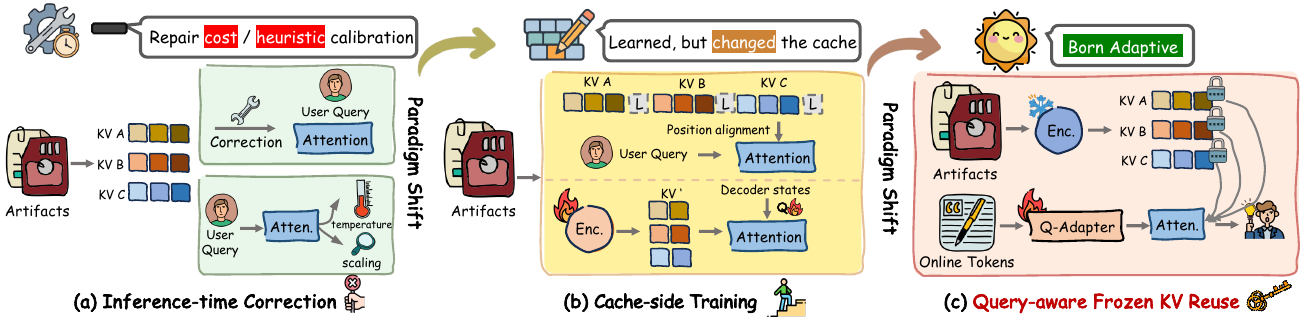}
    \caption{
    \textbf{Three approaches to reusable artifact caching.}
    (a) Inference-time adaptation repairs cached states or adjusts attention during inference.
    (b) Training-based adaptation modifies cache construction through learned encoders or linking tokens.
    (c) \method adapts online queries while preserving independently constructed artifact KV, learning to read the cache without rewriting it.
    }
    \label{fig:paradigm-overview}
\end{figure}

In this work, we examine these mismatches through controlled experiments (\S\ref{sec:observations}). We first find that the positional mismatch has little impact on task performance. To assess whether the cached artifact representations themselves need to be recomputed, we separate the use of artifacts into two abilities: \emph{artifact discovery}, which selects relevant artifacts from the candidates, and \emph{evidence utilization}, which reads and integrates information from the selected artifacts. In the single-artifact setting, evidence utilization suffers little performance loss under PIC, indicating that the cached representation faithfully preserves its own content and remains readable by the model. The failure emerges in the multi-artifact setting: the model struggles to decide which artifact to read from. This contrast localizes the problem to how attention is allocated across artifacts rather than to the cached representations themselves. To test this directly, we replace PIC's online attention scores with prefix-matched full-prefill attention scores while keeping the cached artifact KV unchanged, and find that this recovers performance. These results establish an effective recovery path through \emph{attention routing}, without reconstructing the cached artifact representations.

Motivated by this, we propose \method{}, a query-side adaptation method that learns to read a frozen artifact cache by correcting attention queries, without recomputing or modifying the cached KV.  Figure~\ref{fig:paradigm-overview} contrasts this query-side design with inference-time correction and cache-side training. \method{} inserts low-rank adapters~\citep{hu2022lora} into the query projections, enabling online tokens to learn effective attention routing over cached keys. The adapters are trained through response-distribution distillation, with a full-prefill teacher supervising the PIC student. At inference, no Full-prefill reference or artifact-cache reconstruction is required; online states and newly produced KV continue to evolve under the adapted computation.

We evaluate \method{} on Qwen3-4B and Qwen3-8B across an extensive set of seven benchmarks covering diverse artifact types, measuring task quality and time to first token (TTFT). Further analyses examine training-data scale and show that the learned adaptation brings attention outputs closer to Full-context's on average while leaving cached artifact KV unchanged.

Our key contributions are threefold: (1) We identify the root cause of independent caching degradation: the cached representations are faithful, and the failure lies in the attention score computation. We confirm this with a routing experiment in which replacing PIC's attention scores with full-prefill scores recovers performance while the artifact KV remains unchanged. (2) We propose \method{}, a query-side adaptation method that learns to read a frozen artifact cache through low-rank updates to its query projections. It trains fewer than 0.05\% of the model parameters and, at deployment, requires neither artifact-cache reconstruction nor an additional teacher run. (3) We extensively evaluate \method{} on Qwen3-4B and Qwen3-8B across skill, document and memory tasks. Compared with other PIC baseline, it improves both in-domain quality, out-of-domain generalization and efficiency: average scores rise by 5.13 and 7.77 points, matching the full-context prefill performance while providing up to $3.73\times$ speedups.
\section{Background and Related Work}
\label{sec:pre}

\paragraph{Position-Independent Caching}

An \emph{artifact} $A$ is a context span that can be reused verbatim across requests, such as a skill description, a document, or a memory entry. For exposition, consider a request with a prefix $P$, $m$ artifacts, and a user query $U$:
\begin{equation}
    X = P \oplus A_1 \oplus \cdots \oplus A_m \oplus U,
    \qquad
    Y \sim p_\theta(\cdot \mid X),
\end{equation}
where $\oplus$ denotes concatenation and $Y$ is the generated response.

\par\smallskip
\noindent\fbox{\parbox{\dimexpr\linewidth-2\fboxsep-2\fboxrule\relax}{
\textbf{Full prefill} encodes each artifact under its preceding context in the current request.
\textbf{Position-independent caching (\pic)} instead encodes each artifact independently under a fixed cache-building prefix $P_{\mathrm{cache}}$ and stores the resulting KV states for reuse. At request time, \pic assembles these cached blocks in prompt order without re-encoding the artifacts, computing states only for uncached prompt tokens and subsequently generated tokens. 
}}
\par\smallskip

Formally, let $\mathcal{C}_\theta(A \mid H)$ denote the keys and values across all layers obtained by encoding artifact $A$ after context $H$, and let $A_{<j} = A_1 \oplus \cdots \oplus A_{j-1}$. The two cache constructions are
\begin{equation}
    Z_j^{\mathrm{Full}}
    = \mathcal{C}_\theta\!\left(A_j \mid P \oplus A_{<j}\right),
    \qquad
    Z_j^{\mathrm{PIC}}
    = \mathcal{C}_\theta\!\left(A_j \mid P_{\mathrm{cache}}\right).
    \label{eq:full-vs-pic}
\end{equation}
Consequently, identical artifact text can yield different KV states under Full and \pic.

\paragraph{KV-Cache Reuse beyond Shared Prefixes.}
Prefix caching reuses computation when requests share an identical prefix, as in RadixAttention~\citep{sglang}, while Prompt Cache~\citep{gim2024promptcache} extends this idea to reusable prompt modules with controlled placement. Position-independent reuse is more challenging because cached states are constructed without the request-specific context in which they will later be used. Existing methods largely address this mismatch through \emph{inference-time repair}.
One family selectively restores missing contextual dependencies by recomputing a small portion of the cached context. CacheBlend~\citep{yao2024cacheblend}, Cache-Craft~\citep{cachecraft}, and EPIC~\citep{hu2024epic} recompute selected tokens or artifact boundaries, while later methods use query relevance or learned selection to determine where repair is most useful~\citep{a3,prophetkv,cacheclip,kvshare,redknot}. Rather than recomputing the original states, another family directly corrects their use at inference time. APE~\citep{yang2025ape} calibrates attention over independently encoded blocks, while CacheSlide~\citep{cacheslide} and AgentKVShift~\citep{agentkvshift} introduce position- or probe-based corrections to reused KV states. These approaches highlight two related sources of cache compatibility: \emph{what information is stored in the cache} and \emph{how that information is read after reuse}. The former is also reflected in work showing that prefill can encode intermediate conclusions into downstream states and that cached representations can be edited~\citep{takenotes}.
Complementary works focus on making such reuse practical through cache alignment, recomputation primitives, and serving support~\citep{mepic,minipic}, with extensions to multi-agent and multimodal settings~\citep{kvcomm,tokendance,mpic,vlcache}.

\paragraph{Learning for Cache Reuse.}
A complementary line of work learns cache-compatible behavior rather than repairing mismatches entirely at inference time. Some methods adapt the language model itself to operate over independently encoded blocks. Some methods learn cache-compatible behavior through model adaptation, including block-wise fine-tuning and learned cross-artifact aggregation~\citep{blockattention,turborag,kvlink,recache}. Others instead learn alternative cache representations or interfaces~\citep{encoderpic,c2kv}. Full-context distillation provides another route, training independently constructed caches to approximate full-prefill behavior~\citep{kvpacket,sempic}. Prior learning-based methods primarily modify the backbone, cache construction, or the interface that produces and exposes reusable representations. 
\section{Understanding Quality Loss and Recovery in PIC}
\label{sec:observations}

We first examine how PIC affects task performance across different artifact configurations. Quality loss persists even when relevant artifacts are provided, motivating a closer look at how the model reads independently cached context. 

\begin{figure}[H]
    \centering
    \includegraphics[width=\linewidth]
    {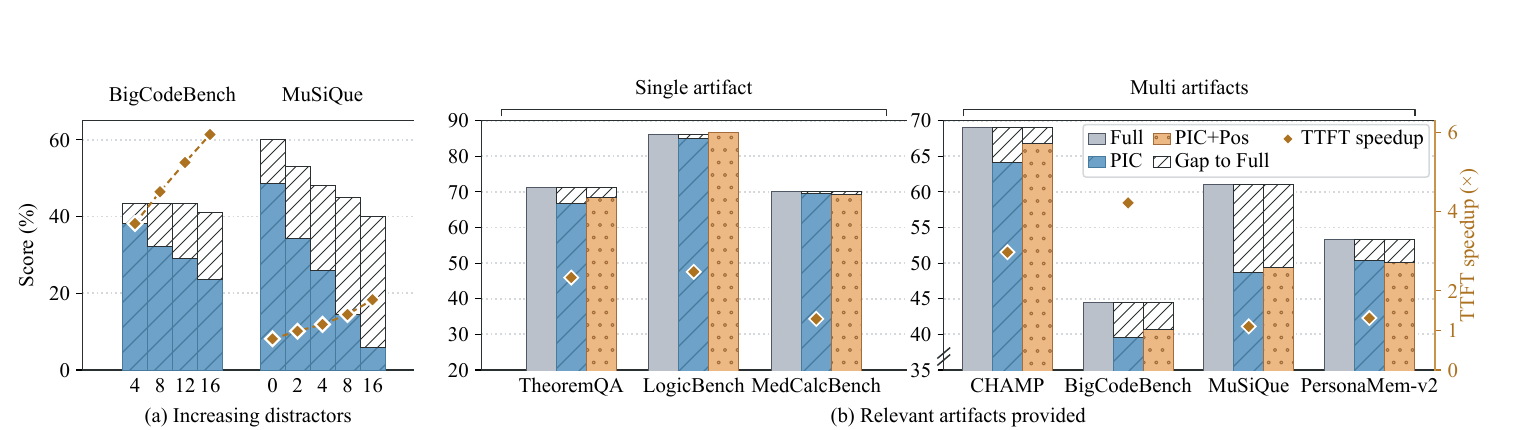}
    \caption{
    Task quality and TTFT speedup on Qwen3-4B.
    (a) Increasing distractors while retaining all relevant skills or supporting documents.
    (b) Supplying only task-annotated relevant artifacts, grouped into single- and multi-artifact settings.
    }
    \label{fig:pic-observations}
\end{figure}

\subsection{Artifact Discovery and Evidence Utilization under PIC}
\label{sec:evidence-utilization}

We begin by examining how PIC impairs the model's use of artifacts when answering questions. Prior PIC systems primarily evaluate generation quality through downstream task scores on multi-chunk inputs~\citep{yao2024cacheblend,hu2024epic,cachecraft}, which combine different aspects of artifact use. Drawing on the distinction between retrieving relevant documents and reading them to answer questions in retrieval-augmented QA~\citep{jiang2022retrieval}, we distinguish \emph{artifact discovery}, which identifies relevant artifacts among the candidates in the prompt, from \emph{evidence utilization}, which reads and integrates information from those artifacts.  We evaluate Qwen3-4B~\citep{qwen3} on SRA-Bench~\citep{su2026skill}, MuSiQue~\citep{trivedi2022musique}, and PersonaMem-v2~\citep{jiang2025personamemv2}, comparing Full and PIC on identical requests under matched generation settings. Experimental details are provided in Section~\ref{sec:experiments}.

\paragraph{Degradation extends beyond artifact discovery.}
Figure~\ref{fig:pic-observations}(a) increases the number of distracting artifacts while retaining all task-annotated relevant skills or supporting documents. PIC deteriorates more sharply than Full on both benchmarks, indicating greater sensitivity to irrelevant context. This pattern is consistent with impaired context-dependent reading when independent cache construction omits cross-artifact conditioning. To examine whether the degradation persists after removing irrelevant candidates, Figure~\ref{fig:pic-observations}(b) compares settings containing only annotated relevant artifacts. PIC incurs relatively small performance losses on the single-artifact tasks, whereas larger gaps remain on multi-artifacts tasks. These results suggest that PIC can retain much of an individual artifact's usefulness while impairing the use of information distributed across multiple artifacts. The degradation that remains \textit{when relevant artifacts are provided} shows that artifact selection alone cannot explain PIC's performance loss. We therefore investigate how the model reads independently cached context to recover effective cache reading while preserving PIC's direct reuse of artifact KV.

\paragraph{Position alignment does not consistently close the gap.}
We next test whether positional mismatch accounts for this remaining degradation. \textbf{PIC-Pos} independently encodes each artifact at its target prompt positions while preserving independent cache construction. This control produces only marginal changes on multi-artifacts task. Position alignment restores positional coordinates but does not recover the conditioning on preceding artifacts that occurs during full prefill. The remaining loss therefore calls for an intervention beyond position correction. 

\subsection{Attention Routing Enables Recovery with Cached Values}
\label{sec:routing-recovery}

Independent caching changes both cached values and attention routing, consistent with prior observations of attention misalignment under parallel encoding~\citep{yang2025ape}. 
We therefore ask: \emph{does the degradation stem from changes in the stored values, the way they are read, or both?}

To examine whether restoring routing is sufficient for recovery, we construct an \emph{attention routing oracle} that retains PIC's cached artifact values.
For each selected online token $t$, it replaces the attention weights with a Full reference and applies them to the intervention branch's own values:
\begin{align}
\Pi_{\ell,h,t}^{\mathrm{Full}}
=
\operatorname{softmax}\!\left(
\frac{q_{\ell,h,t}^{\mathrm{Full}}
(K_{\ell,h,\leq t}^{\mathrm{Full}})^\top}
{\sqrt{d_k}} + M_t
\right),
\qquad
o_{\ell,h,t}^{\mathrm{oracle}}
=
\Pi_{\ell,h,t}^{\mathrm{Full}}
V_{\ell,h,\leq t}^{\mathrm{hybrid}},
\label{eq:routing-oracle}
\end{align}
Here, $M_t$ is the causal attention mask, and $o_{\ell,h,t}^{\mathrm{oracle}}$ is the per-head attention output before the output projection. $V^{\mathrm{hybrid}}$ retains PIC's cached artifact values, while online values evolve under the intervention. We replace complete attention rows at every layer and query head for online prompt tokens following the first artifact and all subsequent decoding tokens. 
\begin{figure}[t]
    \centering
    \includegraphics[width=\linewidth]
    {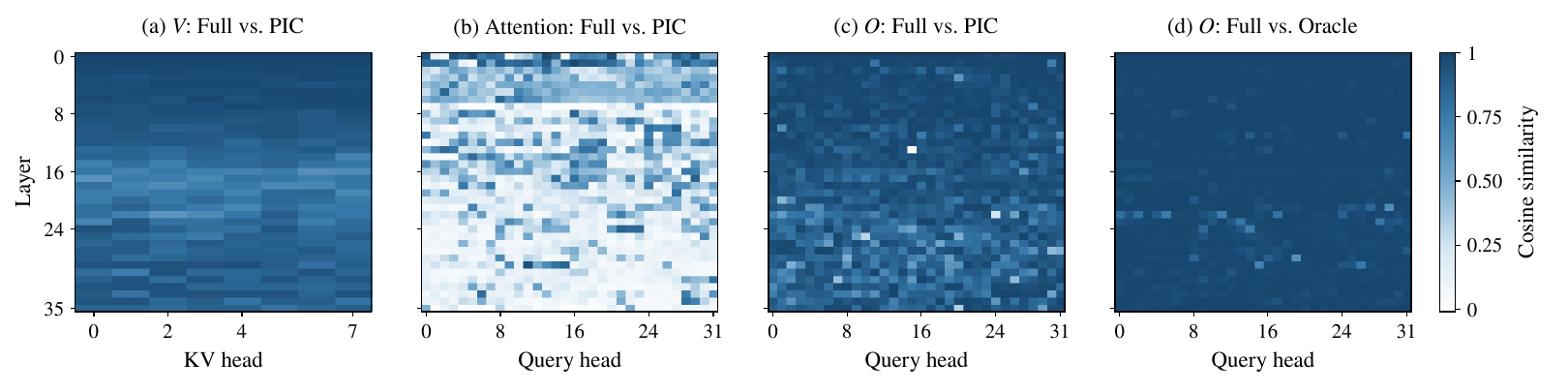}
    \caption{
    \textbf{Routing recovery with unchanged artifact KV.}
    Cosine similarities for one MuSiQue example: Full vs.\ PIC for (a) Full--PIC similarity of cached values across artifact tokens, (b) Full--PIC similarity of attention-weight vectors overartifact tokens, and (c--d) Similarity of per-head attention outputs to Full, under PIC and the routing oracle. Additional case studies appear in Appendix~\ref{app:routing-case-studies}.}
    \label{fig:routing-recovery-cosine}
\end{figure}
Figure~\ref{fig:routing-recovery-cosine} illustrates the effect of this intervention on a paired MuSiQue example.\footnote{Additional repair interventions and their effects on cached representations and attention routing are examined in Appendix~\ref{app:alternative-repairs}.} Cached values remain relatively similar to Full in panel (a), whereas attention allocation over artifact tokens differs substantially in panel (b), particularly in later layers. Because attention outputs combine routing weights with values, routing differences need not translate proportionally into output differences. Replacing routing while retaining artifact KV brings the outputs substantially closer to Full (panels c--d): at the final prompt token, mean output cosine similarity increases from 0.898 to 0.984. Beyond this representation-level example, paired task evaluations show recovery to approximately Full performance. On MuSiQue, EM increases from 34.84 under PIC to \textbf{47.91} under the oracle, compared with \textbf{47.79} under Full. On BigCodeBench, pass@1 increases from 38.98 to \textbf{44.53}, compared with \textbf{43.63} under Full. These paired results establish an effective recovery path: \textit{restoring attention routing can recover task quality while retaining independently cached values.} 
\section{\method: Learning to Read a Frozen Cache}
\label{sec:methodology}

\begin{figure}[h]
    \centering
    \includegraphics[width=\linewidth]{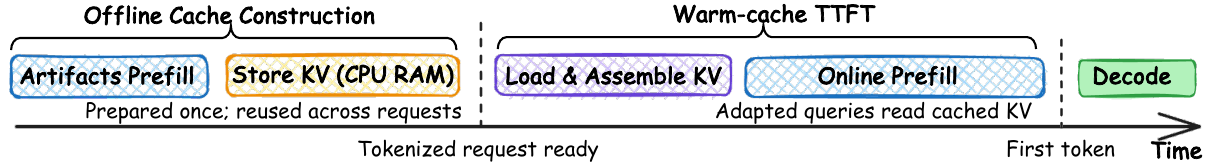}
    \caption{
    \textbf{Inference workflow of \method.}
    Artifact KV is prepared once and reused unchanged. Warm-cache TTFT includes KV loading and assembly, online prefill, and first-token selection, excluding offline cache construction.
    }
    \label{fig:attuner-overview}
\end{figure}

Motivated by the cache-preserving recovery path in Section~\ref{sec:routing-recovery}, We introduce \method, a simple yet efficient method to read independently cached artifacts. As illustrated in Figure~\ref{fig:attuner-overview}, artifacts are prefilled in advance and reused across requests, while adapted queries operate during online prefill and decoding. We explain why query-side adaptation preserves cache reuse, then describe its training through Full-context supervision.

\paragraph{Query Adaptation with Fixed Artifact KV}
\label{sec:query-adaptation}

Although both queries and keys determine attention weights, adapting queries preserves the existing artifact cache: \textit{queries are computed for online tokens, while cached keys and values remain available for direct reuse}. We augment the frozen query projection in every Transformer layer with a low-rank update~\citep{hu2022lora}:
\begin{equation}
\widetilde{q}_{\ell,t}
=
W_{\ell}^{Q} h_{\ell,t}
+
\frac{\alpha}{r} B_{\ell} A_{\ell} h_{\ell,t},
\qquad \ell = 1,\ldots,L,
\label{eq:query-adaptation}
\end{equation}
where $h_{\ell,t}$ is the projection input, $A_{\ell} \in \mathbb{R}^{r \times d_{\mathrm{model}}}$ and $B_{\ell} \in \mathbb{R}^{d_q \times r}$ are trainable matrices, $r$ is the adapter rank, and $\alpha$ controls the update scale. The adapted projection $\widetilde{q}_{\ell,t}$ then passes through the model's original query normalization and RoPE before entering attention. We initialize the update to zero and train only $A_{\ell}$ and $B_{\ell}$, keeping all base-model parameters frozen.
By enabling adapters only for uncached prompt and generated tokens, \method adapts cache reading while preserving artifact KV construction and reuse. \textit{Stored artifact KV remains unchanged}; online states and newly produced KV \textit{evolve under the adapted computation}. 

\paragraph{Learning from Full-Context Behavior}
\label{sec:training-obj}

\begin{figure}[t]
    \centering
    \includegraphics[width=\linewidth]
    {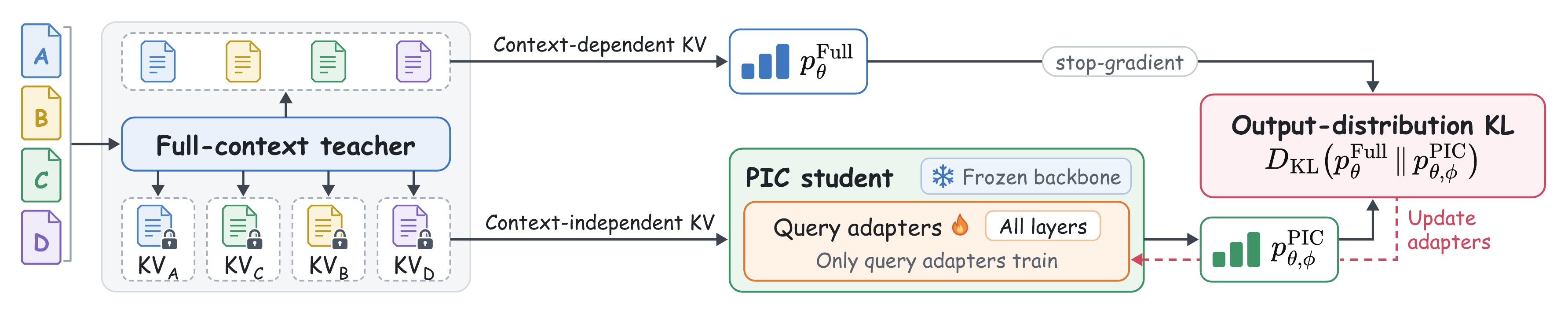}
    \caption{
    \textbf{Training \method.} A frozen Full-context teacher supervises the PIC student's output distributions under identical inputs and response prefixes. Only query adapters are updated; the backbone and cached
artifact KV remain frozen.
    }
    \label{fig:attuner-training}
\end{figure}

As shown in Figure~\ref{fig:attuner-training}, we distill Full-context answering behavior into the PIC student. For each request $X$, the frozen teacher generates a response $Y$, forming a fixed training set $\mathcal{D}_T$. Both branches receive the same input and condition on the same response prefix $Y_{<t}$, but the teacher uses full prefill while the student reuses independently cached artifact KV. We minimize
\begin{equation}
\mathcal{L}(\phi)
=
\mathbb{E}_{(X,Y)\sim\mathcal{D}_T}
\left[
\frac{1}{|Y|}
\sum_{t=1}^{|Y|}
D_{\mathrm{KL}}\!\left(
\operatorname{sg}\!\left[
p_{\theta}^{\mathrm{Full}}(\cdot\mid X,Y_{<t})
\right]
\,\middle\|\,
p_{\theta,\phi}^{\mathrm{PIC}}(\cdot\mid X,Y_{<t})
\right)
\right],
\label{eq:offpolicy-kl}
\end{equation}
where $\phi=\{A_{\ell},B_{\ell}\}_{\ell=1}^{L}$ denotes the query-adapter parameters and $\operatorname{sg}$ stops gradients through the teacher. Supervising output distributions rather than attention weights allows \method to learn a reading strategy suited to the fixed cache. At deployment, the trained updates are merged into the query projections, requiring no Full-context teacher.
\section{Experiments}
\label{sec:experiments}

\subsection{Experimental Setup}
\label{sec:experimental-setup}

\paragraph{Implementation details.}
We evaluate \method on Qwen3-4B and Qwen3-8B~\citep{qwen3} using skills from SRA-Bench~\citep{su2026skill}, documents from MuSiQue~\citep{trivedi2022musique}, and memory messages from PersonaMem-v2~\citep{jiang2025personamemv2}. SRA-Bench includes TheoremQA~\citep{chen2023theoremqa}, LogicBench~\citep{parmar2024logicbench}, MedCalc-Bench~\citep{khandekar2024medcalc}, CHAMP~\citep{mao2024champ}, and BigCodeBench~\citep{zhuo2025bigcodebench} (excluding ToolQA~\citep{zhuang2023toolqa}, which reuses a single skill across multi-step ReAct interactions). For each base model, we train one shared adapter on a balanced mixture of 12,000 examples: 4,000 each from the training splits of MuSiQue and PersonaMem-v2, and 4,000 synthetic skill-use examples constructed from BigCodeBench training-side skills. These skills are disjoint from those used for evaluation. Full-context responses from the corresponding base model provide supervision, and the same adapter is evaluated across all tasks.

\paragraph{Baselines.}
We compare \method with Full prefill and unmodified PIC. Our baselines include training-free methods APE~\citep{yang2025ape}, EPIC~\citep{hu2024epic}, CacheBlend~\citep{yao2024cacheblend}, and AgentKVShift~\citep{agentkvshift}, as well as training-based methods KVLink~\citep{kvlink} and COMB~\citep{encoderpic}. Method-specific cache construction, training, and implementation details are provided in Appendix~\ref{app:baselines}.

\paragraph{Evaluation metrics.}
We report pass@1 for BigCodeBench, F1 for MuSiQue, and accuracy for the remaining tasks. For efficiency, we measure median warm-cache time to first token (TTFT), with reusable KV already resident in CPU memory. Timing includes cache transfer to the GPU and assembly, method-specific repair, uncached-token prefill, and first-token selection; it excludes model loading, tokenization, and offline cache construction. Speedup is Full's median TTFT divided by that of the compared method under matched examples, hardware, and inference settings.

\subsection{Main Results}
\label{sec:main-results}
\begin{figure*}[t]
\centering

\begin{minipage}[t]{0.46\textwidth}
    \vspace{0pt}
    \centering

    \includegraphics[width=\linewidth]
    {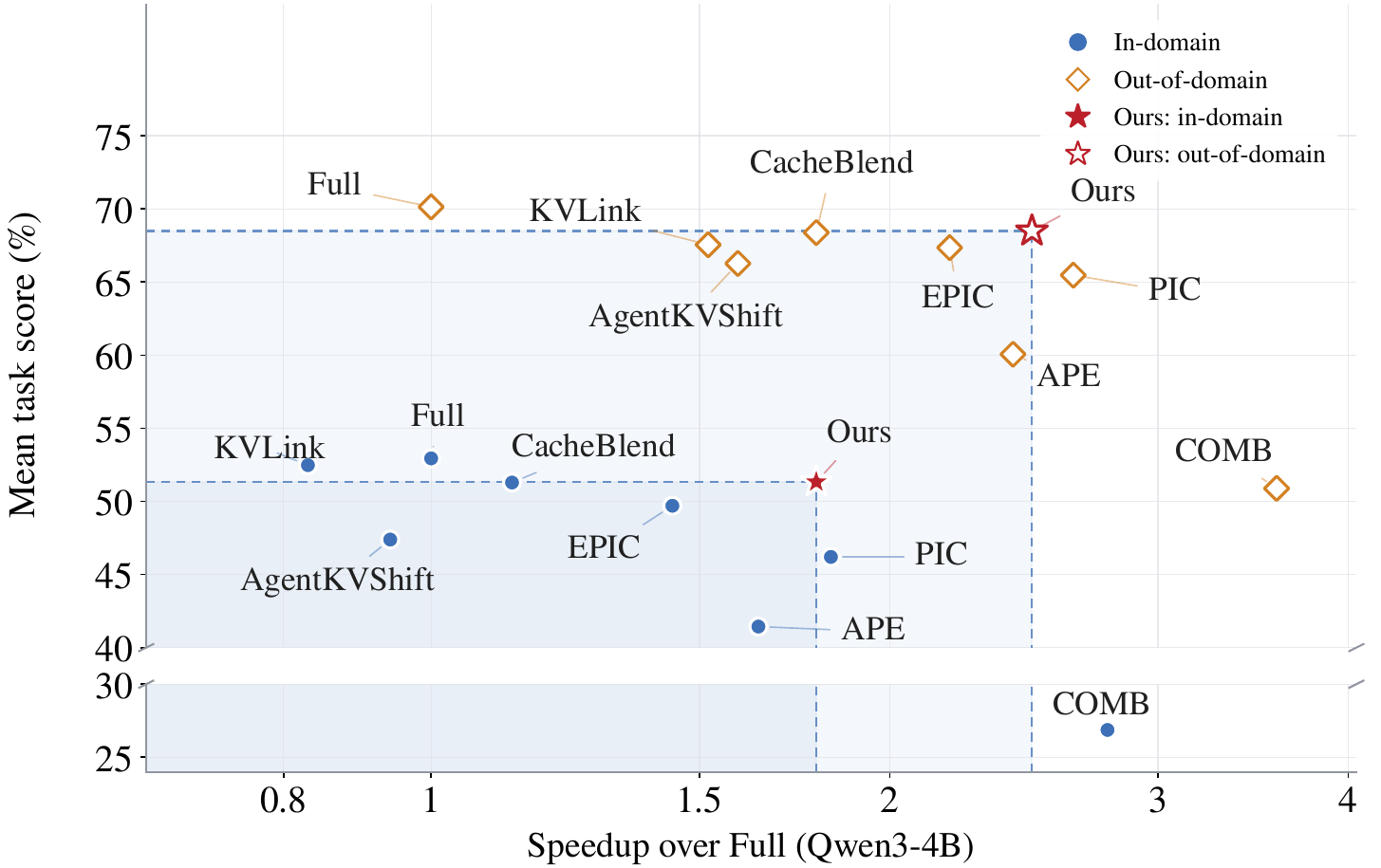}

    \captionof{figure}{
    \textbf{Quality--efficiency trade-off.}
    Quality versus speedup.
    Shaded regions indicate lower quality and speed than \method.
    }
    \label{fig:quality-efficiency}
\end{minipage}
\hfill
\begin{minipage}[t]{0.52\textwidth}
    \vspace{0pt}
    \centering
    \captionof{table}{
    \textbf{Single-artifact quality and efficiency.}
    TQA, LB, and MCB denote TheoremQA, LogicBench, and
    MedCalc-Bench, respectively.
    }
    \begingroup
    \setlength{\tabcolsep}{2.6pt}
    \renewcommand{\arraystretch}{1.12}
    \scriptsize

    \begingroup

\fontsize{9}{11}\selectfont

\setlength{\tabcolsep}{3pt}
\renewcommand{\arraystretch}{1.5}

\providecommand{\scoredown}[2]{}
\renewcommand{\scoredown}[2]{%
  #1~{\scriptsize\textcolor{green!60!black}{$\downarrow$\,#2}}%
}
\providecommand{\scoreup}[2]{}
\renewcommand{\scoreup}[2]{%
  #1~{\scriptsize\textcolor{red!70!black}{$\uparrow$\,#2}}%
}
\providecommand{\ttftspeed}[2]{}
\renewcommand{\ttftspeed}[2]{%
  #1~{\scriptsize($#2\times$)}%
}

\resizebox{\linewidth}{!}{%
\begin{NiceTabular}{
    >{\centering\arraybackslash}m{10mm}
    c
    *{5}{c}
}
\toprule

\Block{2-1}{}
& \Block{2-1}{{\scriptsize Dataset}}
& \multicolumn{3}{c}{\scriptsize\textbf{Score}}
& \multicolumn{2}{c}{\scriptsize\textbf{Efficiency}} \\
\cmidrule(lr){3-5}
\cmidrule(lr){6-7}

&
& {\scriptsize Full}
& {\scriptsize PIC}
& {\scriptsize PIC-Pos}
& {\scriptsize Full}
& {\scriptsize PIC / PIC-Pos} \\

\midrule

\Block{3-1}{
  \rotatebox[origin=c]{90}{%
    {\scriptsize\bfseries Qwen3-4B}%
  }
}
& {\scriptsize TQA}
& 71.22
& \scoredown{66.80}{4.42}
& \scoredown{68.54}{2.68}
& 89.08
& \ttftspeed{38.14}{2.34} \\

& {\scriptsize LB}
& 86.05
& \scoredown{85.00}{1.05}
& \scoreup{86.71}{0.66}
& 94.43
& \ttftspeed{38.08}{2.48} \\

& {\scriptsize MCB}
& 70.09
& \scoredown{69.55}{0.54}
& \scoredown{69.27}{0.82}
& 124.61
& \ttftspeed{96.50}{1.29} \\

\specialrule{0.6pt}{0pt}{0pt}

\Block{3-1}{
  \rotatebox[origin=c]{90}{%
    {\scriptsize\bfseries Qwen3-8B}%
  }
}
& {\scriptsize TQA}
& 63.05
& \scoredown{55.15}{7.90}
& \scoredown{52.34}{10.71}
& 145.85
& \ttftspeed{38.89}{3.75} \\

& {\scriptsize LB}
& 87.24
& \scoredown{85.79}{1.45}
& \scoreup{87.63}{0.39}
& 156.91
& \ttftspeed{40.15}{3.91} \\

& {\scriptsize MCB}
& 72.91
& \scoredown{72.27}{0.64}
& \scoredown{72.27}{0.64}
& 207.36
& \ttftspeed{140.12}{1.48} \\

\bottomrule
\end{NiceTabular}%
}

\endgroup

    \endgroup

    \label{tab:single-artifact-results}
\end{minipage}

\end{figure*}

\paragraph{Single-artifact caching offers substantial speedups, but uneven quality.}
Table~\ref{tab:single-artifact-results} shows that direct cache reuse achieves 1.29--3.91$\times$ speedup over Full. On LogicBench and MedCalc-Bench, PIC remains within 1.45 accuracy points of Full across both model sizes, providing an attractive quality--latency trade-off. TheoremQA exposes a limitation: accuracy drops by 4.42 points on Qwen3-4B and 7.90 points on Qwen3-8B, and \textit{position alignment does not consistently recover the loss}. We therefore also include TheoremQA as an out-of-distribution task to evaluate whether \method's learned cache-reading behavior transfers beyond the training benchmarks.

\newcommand{\deltadown}[1]{%
  {\textcolor{green!60!black}{$\downarrow$\,#1}}%
}

\newcommand{\deltaup}[1]{%
  {\textcolor{red!70!black}{$\uparrow$\,#1}}%
}

\begin{table*}[t]
\centering

\caption{
\textbf{Quality and efficiency on training-domain and held-out tasks.} BCB, MSQ, PMem, TQA, and CHP denote BigCodeBench, MuSiQue, PersonaMem, TheoremQA, and CHAMP, respectively. Avg is the arithmetic mean of scores within each domain. Geo. denotes the geometric mean of TTFT within each domain. Full and PIC are the references and excluded from ranking. The best and second-best results among the remaining methods are \textbf{bold} and \underline{underlined}.
}
\begingroup
\setlength{\tabcolsep}{1.8pt}
\renewcommand{\arraystretch}{1.10}

\resizebox{\textwidth}{!}{%
\begin{NiceTabular}{cl*{18}{c}}
\CodeBefore
    \rectanglecolor{black!6}{6-2}{6-20}
    \rectanglecolor{black!6}{11-2}{11-20}
    \rectanglecolor{black!6}{17-2}{17-20}
    \rectanglecolor{black!6}{22-2}{22-20}
    \rectanglecolor{green!10}{14-2}{14-20}
    \rectanglecolor{green!10}{25-2}{25-20}
\Body

\toprule

\Block{3-1}{}
&
& \multicolumn{10}{c}{\textbf{In-domain}}
& \multicolumn{8}{c}{\textbf{Out-of-domain}} \\
\cmidrule(lr){3-12}
\cmidrule(lr){13-20}

& Method
& \multicolumn{2}{c}{BCB}
& \multicolumn{2}{c}{MSQ}
& \multicolumn{2}{c}{PMem}
& \multicolumn{2}{c}{Quality}
& \multicolumn{2}{c}{Efficiency}
& \multicolumn{2}{c}{TQA}
& \multicolumn{2}{c}{CHP}
& \multicolumn{2}{c}{Quality}
& \multicolumn{2}{c}{Efficiency} \\

&
& Score $\uparrow$ & TTFT $\downarrow$
& Score $\uparrow$ & TTFT $\downarrow$
& Score $\uparrow$ & TTFT $\downarrow$
& Avg $\uparrow$ & $\Delta$
& Geo. $\downarrow$ & Speedup $\uparrow$
& Score $\uparrow$ & TTFT $\downarrow$
& Score $\uparrow$ & TTFT $\downarrow$
& Avg $\uparrow$ & $\Delta$
& Geo. $\downarrow$ & Speedup $\uparrow$ \\

\midrule

\Block{11-1}{\rotatebox{90}{\textbf{Qwen3-4B}}}
& Full
& 44.53 & 420.77
& 61.01 & 39.06
& 53.28 & 62.38
& 52.94 & --
& 100.83 & $1.00\times$
& 71.22 & 89.08
& 69.06 & 139.01
& 70.14 & --
& 111.28 & $1.00\times$ \\

& PIC
& 39.58 & 99.60
& 48.69 & 35.51
& 50.34 & 47.56
& 46.20 & \deltadown{6.74}
& 55.20 & $1.83\times$
& 66.80 & 38.14
& 64.13 & 46.70
& 65.47 & \deltadown{4.68}
& 42.20 & $2.64\times$ \\

& \multicolumn{19}{c}{\textit{\textbf{Training-free}}} \\

& APE
& 39.58 & 114.59
& 41.02 & 39.17
& 43.72 & 51.92
& 41.44 & \deltadown{11.50}
& 61.54 & $1.64\times$
& 59.57 & 42.88
& 60.54 & 49.83
& 60.06 & \deltadown{10.09}
& 46.22 & $2.41\times$ \\

& EPIC
& 44.08 & \underline{100.57}
& 54.21 & 50.76
& 50.80 & 67.29
& 49.70 & \deltadown{3.24}
& 70.03 & $1.44\times$
& 67.87 & 46.73
& 66.82 & 55.23
& 67.35 & \deltadown{2.80}
& 50.80 & $2.19\times$ \\

& CacheBlend
& \textbf{45.13} & 194.81
& 54.82 & 52.43
& \textbf{53.92} & 69.73
& 51.29 & \deltadown{1.65}
& 89.30 & $1.13\times$
& \underline{68.14} & 50.95
& \textbf{68.61} & 75.92
& \underline{68.38} & \deltadown{1.77}
& 62.19 & $1.79\times$ \\

& \small{AgentKVShift}
& 42.13 & 196.54
& 52.62 & 68.78
& 47.42 & 90.25
& 47.39 & \deltadown{5.55}
& 106.85 & $0.94\times$
& 67.07 & 61.59
& 65.47 & 79.94
& 66.27 & \deltadown{3.87}
& 70.17 & $1.59\times$ \\

& \multicolumn{19}{c}{\textit{\textbf{Training-based}}} \\

& KVLink
& \underline{44.98} & 165.04
& \textbf{59.93} & 98.70
& \underline{52.54} & 109.94
& \textbf{52.48} & \deltadown{0.46}
& 121.44 & $0.83\times$
& 66.93 & 68.96
& \underline{68.16} & 78.02
& 67.55 & \deltadown{2.60}
& 73.35 & $1.52\times$ \\

& COMB
& 35.68 & \textbf{38.29}
& 12.70 & \textbf{28.68}
& 32.22 & \textbf{43.49}
& 26.87 & \deltadown{26.07}
& \textbf{36.28} & $\mathbf{2.78\times}$
& 47.52 & \textbf{30.78}
& 54.26 & \textbf{31.18}
& 50.89 & \deltadown{19.25}
& \textbf{30.98} & $\mathbf{3.59\times}$ \\

& \textbf{\method}
& 44.08 & 101.27
& \underline{58.06} & \underline{36.71}
& 51.86 & \underline{47.75}
& \underline{51.33} & \deltadown{1.61}
& \underline{56.20} & $\underline{1.79\times}$
& \textbf{68.81} & \underline{39.80}
& \underline{68.16} & 50.62
& \textbf{68.49} & \deltadown{1.66}
& \underline{44.89} & $\underline{2.48\times}$ \\

\specialrule{1.1pt}{0pt}{0pt}

\Block{11-1}{\rotatebox{90}{\textbf{Qwen3-8B}}}
& Full
& 50.22 & 660.41
& 63.28 & 62.89
& 43.90 & 99.14
& 52.47 & --
& 160.28 & $1.00\times$
& 63.05 & 145.85
& 62.78 & 227.68
& 62.92 & --
& 182.23 & $1.00\times$ \\

& PIC
& 47.08 & 113.82
& 50.00 & 35.33
& 35.50 & 68.60
& 44.19 & \deltadown{8.27}
& 65.10 & $2.46\times$
& 55.15 & 38.89
& 63.23 & 59.17
& 59.19 & \deltadown{3.73}
& 47.97 & $3.80\times$ \\

& \multicolumn{19}{c}{\textit{\textbf{Training-free}}} \\

& APE
& 46.63 & 123.16
& 44.94 & 39.60
& \underline{44.86} & 73.10
& 45.48 & \deltadown{6.99}
& 70.91 & $2.26\times$
& 52.34 & 45.10
& 51.12 & 60.80
& 51.73 & \deltadown{11.19}
& 52.36 & $3.48\times$ \\

& EPIC
& 48.58 & \underline{110.50}
& 56.71 & 50.79
& 42.40 & 88.41
& 49.23 & \deltadown{3.24}
& 79.17 & $2.02\times$
& 58.23 & 58.89
& 61.88 & 67.53
& 60.06 & \deltadown{2.86}
& 63.06 & $2.89\times$ \\

& CacheBlend
& 48.43 & 249.41
& 58.12 & 54.92
& 42.78 & 96.36
& 49.78 & \deltadown{2.69}
& 109.69 & $1.46\times$
& 57.56 & 66.42
& 42.60 & 101.32
& 50.08 & \deltadown{12.84}
& 82.03 & $2.22\times$ \\

& \small{AgentKVShift}
& 47.53 & 246.76
& 55.58 & 68.88
& 40.92 & 106.07
& 48.01 & \deltadown{4.46}
& 121.71 & $1.32\times$
& \textbf{61.58} & 80.25
& 64.13 & 105.39
& \textbf{62.86} & \deltadown{0.06}
& 91.96 & $1.98\times$ \\

& \multicolumn{19}{c}{\textit{\textbf{Training-based}}} \\

& KVLink
& \underline{50.22} & 182.91
& \underline{61.20} & 102.70
& \textbf{46.56} & 143.02
& \textbf{52.66} & \deltaup{0.19}
& 139.02 & $1.15\times$
& 58.50 & 70.52
& \textbf{66.37} & 92.46
& 62.44 & \deltadown{0.48}
& 80.75 & $2.26\times$ \\

& COMB
& 42.13 & \textbf{48.17}
& 14.90 & \textbf{31.02}
& 31.62 & \textbf{67.51}
& 29.55 & \deltadown{22.92}
& \textbf{46.55} & $\mathbf{3.44\times}$
& 42.84 & \underline{42.24}
& 51.57 & \textbf{44.33}
& 47.21 & \deltadown{15.71}
& \textbf{43.27} & $\mathbf{4.21\times}$ \\

& \textbf{\method}
& \textbf{50.52} & 113.19
& \textbf{61.22} & \underline{36.33}
& 44.14 & \underline{68.10}
& \underline{51.96} & \deltadown{0.51}
& \underline{65.42} & $\underline{2.45\times}$
& \underline{60.37} & \textbf{39.80}
& \underline{65.02} & \underline{59.98}
& \underline{62.70} & \deltadown{0.22}
& \underline{48.86} & $\underline{3.73\times}$ \\

\bottomrule
\end{NiceTabular}%
}

\endgroup

\label{tab:multi-artifact-results}
\end{table*}

\paragraph{\method recovers quality within and beyond the training tasks.}
Table~\ref{tab:multi-artifact-results} shows that \method improves over PIC on all five tasks for both model sizes. For readability, Appendix~\ref{app:domain-results} presents enlarged tables of the in-domain and out-of-domain results separately. Across the three in-domain benchmarks, average scores increase by 5.13 and 7.77 points on Qwen3-4B and Qwen3-8B, reducing the gaps to Full to 1.61 and 0.51 points, respectively.
\textbf{A single adapter improves the use of skills, documents, and memories while preserving their original cached KV.} Among the repair methods\footnote{Baseline hyperparameters and implementation details are provided in Appendix~\ref{app:baselines}.}, \method achieves the second-highest in-domain average on both models, behind KVLink; on Qwen3-8B, it obtains the highest scores on BigCodeBench and MuSiQue. The benefits also extend to TheoremQA and CHAMP, \textit{neither of which is included in adapter training}: their average scores improve over PIC by 3.02 and 3.51 points. These results demonstrate transfer beyond the training tasks, although recovery remains incomplete on some benchmarks.

\paragraph{\method recovers quality at near-PIC latency.}
With query updates merged into the projection weights, \method's geometric-mean in-domain TTFT is only 1.8\% and 0.5\% higher than PIC on Qwen3-4B and Qwen3-8B, preserving 1.79$\times$ and 2.45$\times$ speedups over Full. Unlike recomputation-based repair, \method retains the original artifact KV and requires \textit{neither online token selection nor artifact-token recomputation}.
Figure~\ref{fig:quality-efficiency} illustrates the resulting trade-off: \method attains \textbf{higher average quality and lower TTFT than all evaluated training-free repair baselines}. KVLink achieves higher average quality, but requires more than twice \method's geometric-mean TTFT on both models. COMB is faster, but incurs substantially larger quality losses under our evaluation setting. Thus, these results show that adapting cache reading can recover quality while retaining the latency advantages of direct cache reuse.

\Needspace{0.40\textheight}
\subsection{Analysis}
\label{sec:in-depth-analysis}
\label{sec:ablations}

\begin{wrapfigure}{r}{0.4\linewidth}
    \centering
    \vspace{-5mm}
    \includegraphics[width=\linewidth]
    {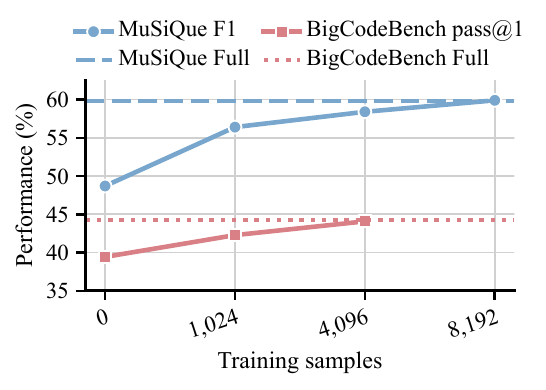}
    \caption{
    Effect of training data size on Qwen3-4B for 1 epoch. Zero examples denotes PIC without adaptation;
    }
    \label{fig:training-data-scale}
    \vspace{-6mm}
\end{wrapfigure}

\paragraph{More training data progressively closes the quality gap.} Figure~\ref{fig:training-data-scale} shows that \textbf{limited training data already enables substantial recovery}. On MuSiQue, 1,024 examples raise F1 from 48.70 to 56.40; 8,192 examples bring it to 61.15, matching the Full baseline of 61.01. BigCodeBench similarly improves from 39.43 to 44.08 pass@1, approaching Full's 44.53. These results show that substantial recovery is possible with limited training data, with further gains as training coverage expands.

\begin{figure*}[ht]
    \centering
    \includegraphics[width=\textwidth]
    {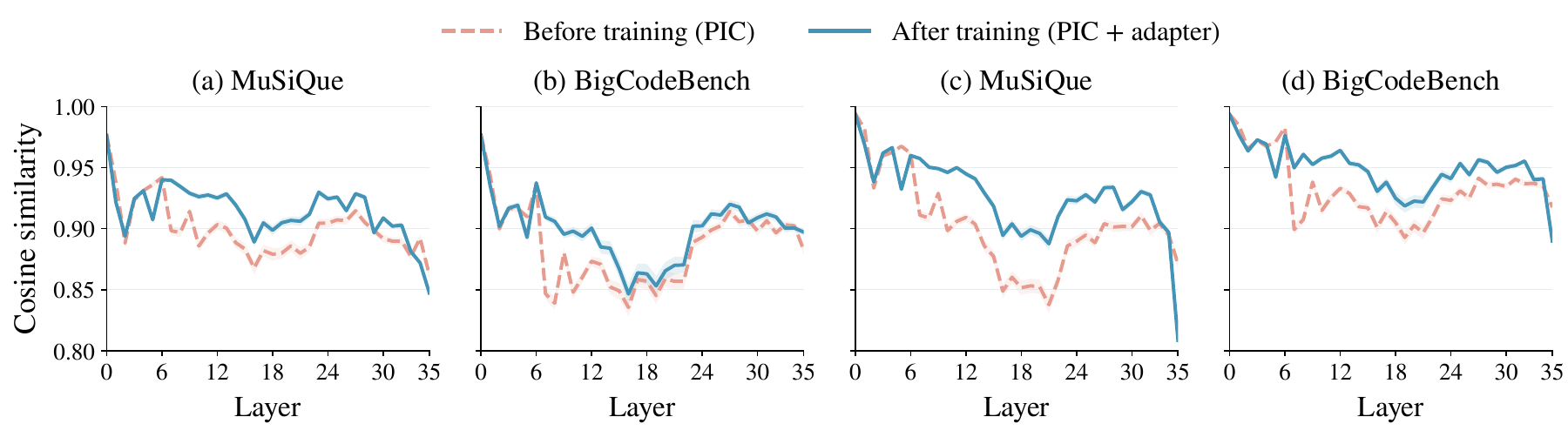}

    \caption{
    \textbf{Attention-output alignment before and after training.} Cosine similarity of $O$ to Full on Qwen3-4B. The first two panels show online prompt positions; the last two show teacher-forced answer positions. Each pair presents MuSiQue followed by BigCodeBench, with 128 examples.
    }
    \label{fig:training-output-alignment}
\end{figure*}

\paragraph{Query adaptation improves attention-output alignment.} 
We compare per-head attention outputs $O$ before the output projection, using identical prompts and teacher-forced Full-response prefixes. Figure~\ref{fig:training-output-alignment} shows improved alignment with Full across many intermediate layers, at both online prompt and answer positions. During the answer phase, mean cosine similarity increases from 0.903 to 0.927 on MuSiQue and from 0.932 to 0.949 on BigCodeBench.
The improvement is not uniform: alignment deteriorates at the final layer. Importantly, \method is trained on output distributions, without directly supervising attention weights or $O$. The resulting alignment provides \textit{representation-level evidence that behavioral distillation improves how the fixed cache is used}, without requiring exact reproduction of Full's layer-wise computation.

\paragraph{Adapting a subset of layers can recover quality.}
The uneven alignment changes across layers raise a related question: does effective cache reading require adapting every layer? We examine this through two equally sized adaptation windows, L16--33 and L18--35, each covering half of Qwen3-4B's layers. Both use 1,024 MuSiQue training examples and a shared teacher under matched training settings. On 128 test samples, they achieve 65.14 and 63.34 F1, respectively, compared with 49.00 for PIC (Table~\ref{tab:full-adapter-control}). Although L16--33 scores higher, the difference is inconclusive under paired bootstrap resampling. Together with the preceding alignment analysis, these results show that recovery need not reproduce Full's computation at every layer: substantial gains remain possible when trainable updates are confined to a subset of layers.

\Needspace{0.50\textheight}
\begin{wraptable}[22]{r}{0.48\textwidth}
\centering
\vspace{-3.5mm}
\caption{
\textbf{Adapter controls on Qwen3-4B.}
(a) Layer-range comparison on 128 MuSiQue test examples after training on 1,024 examples
(b) The same L0--35 adapter under Full and PIC; The two panels use different training and evaluation sets.
}
\begingroup
\small
\setlength{\tabcolsep}{4pt}
\renewcommand{\arraystretch}{1.10}

\resizebox{\linewidth}{!}{%
\begin{tabular}{lrrrr}
\toprule
\multicolumn{5}{c}{\textit{(a) Layer-range comparison on MuSiQue}} \\
\midrule
Method & Layers & EM & F1 & $\Delta$ F1 \\
\midrule
Full
& -- & 46.09 & 60.06 & -- \\
PIC
& -- & 34.38 & 49.00 & -- \\
PIC + \method
& 16--33 & 50.78 & 65.14 & $+16.13$ \\
PIC + \method
& 18--35 & 50.00 & 63.34 & $+14.34$ \\

\midrule
\multicolumn{5}{c}{\textit{(b) Full versus PIC adaptation}} \\
\midrule
& \multicolumn{3}{c}{In-domain} & Held-out \\
\cmidrule(lr){2-4}
\cmidrule(lr){5-5}
Method & BCB & MSQ & PMem & CHP \\
\midrule
Full
& 44.53 & 61.01 & 53.28 & 69.06 \\
Full + \method
& 41.98 & 63.70 & 53.64 & 64.57 \\
$\Delta$
& $-2.55$ & $+2.69$ & $+0.36$ & $-4.49$ \\
\midrule
PIC
& 39.58 & 48.69 & 50.34 & 64.13 \\
PIC + \method
& 44.08 & 58.06 & 51.86 & 68.16 \\
$\Delta$
& $+4.50$ & $+9.37$ & $+1.52$ & $+4.03$ \\
\bottomrule
\end{tabular}%
}
\endgroup
\label{tab:full-adapter-control}

\end{wraptable}

\paragraph{The gains are larger under PIC than under full prefill.}
We apply the same trained Qwen3-4B adapter to Full and PIC to assess whether its benefits reflect a general improvement in task performance. As shown in Table~\ref{tab:full-adapter-control}, adaptation improves PIC on all four tasks, whereas its effect under Full is mixed: MuSiQue and PersonaMem-v2 improve, but BigCodeBench and CHAMP decline. Across the three in-domain tasks, the average gain is 5.13 points under PIC, compared with only 0.17 points under Full. Moreover, \textbf{the adaptation gain is larger under PIC on every task}. This contrast supports adaptation to independently cached context as the primary source of improvement, rather than a uniform benefit from additional training.
\section{Conclusion}
\label{sec:conclusion}
We show that PIC's quality loss extends beyond artifact discovery to multi-artifact evidence utilization, and that positional alignment alone does not consistently recover performance. Restoring full-prefill attention routing recovers task quality with cached artifact KV unchanged, establishing a recovery path without cache reconstruction. Motivated by this finding, \method{} learns to read a frozen artifact cache through low-rank query updates trained by distilling full-prefill output distributions. It trains fewer than 0.05\% of model parameters and requires neither artifact-cache recomputation nor a teacher at inference. Across skill, document, memory, and code tasks, a single adapter per model improves average in-domain scores over PIC by 5.13 and 7.77 points on Qwen3-4B and Qwen3-8B, respectively, and transfers to held-out tasks. These gains come at near-PIC warm-cache TTFT, retaining in-domain geometric-mean speedups of 1.79$\times$ and 2.45$\times$ over full prefill. These results highlight query-side adaptation as an effective path toward high-quality, efficient cache reuse: learning to read the cache rather than rewriting it. Limitations and future directions are discussed in Appendix~\ref{app:limitations}.

\subsection*{AI use statement}

Generative AI tools were used to polish the manuscript, assist in finding relevant research, and support codedevelopment and figure preparation. The authors take responsibility for the final text, citation accuracy, implementation correctness, and scientific claims, including all AI-assisted content.

\subsection*{Ethics statement}

This work studies efficient inference using existing models and benchmarks. The proposed method does not address biases, privacy risks, or harmful outputs inherited from the underlying models and data. Applications involving reusable context should ensure appropriate access controls and protection of cached information.

\subsection*{Reproducibility statement}

Section~\ref{sec:methodology} describes the adaptation mechanism and training objective. Section~\ref{sec:experimental-setup} specifies the benchmarks, training data mixture, evaluation metrics, and TTFT measurement protocol. Data preparation, training settings, hardware, and evaluation protocols are provided in Appendix~\ref{app:details}. Baseline implementations and configurations are detailed in Appendix~\ref{app:baselines}.

\normalem

\bibliography{references}

\clearpage

\appendix
\section*{Appendix}
\section{Experimental Details}
\label{app:details}

This appendix describes the data, prompt construction, and training and generation settings used in the main experiments. The diagnostic studies and data-scale ablations use their own explicitly specified cohorts; their reference scores should not be interchanged with main-table results.

\subsection{Datasets and Splits}
\label{app:datasets}

We evaluate Qwen3-4B and Qwen3-8B~\citep{qwen3} on seven benchmarks covering reusable skills, documents, and memory messages. Table~\ref{tab:app-dataset-settings} summarizes the prepared evaluation sets. In the relevant-only setting, all annotated relevant artifacts are supplied directly, without a retrieval stage or additional distractors. The evaluated requests and artifact order are matched across methods. Appendix~\ref{app:dataset-examples} illustrates the cached artifacts, online questions, and evaluation targets with one prepared example each from BigCodeBench, MuSiQue, and PersonaMem-v2.

\begin{table}[!htbp]
    \centering
    \small
    \setlength{\tabcolsep}{0.6em}
    \renewcommand{\arraystretch}{1.12}
    \caption{\textbf{Main evaluation sets.} Artifact counts are per request. Training counts refer to the shared 12K mixture; BigCodeBench training examples are synthetic. MuSiQue uses its official development set for held-out evaluation.}
    \label{tab:app-dataset-settings}
    \vspace{2pt}
    \begin{NiceTabular}{@{}X[l]lrrrl@{}}[width=\linewidth]
\CodeBefore
    \rowcolor{black!6}{1}
\Body
        \toprule
        \textbf{Dataset} & \textbf{Artifact unit} & \textbf{Train} & \textbf{Eval.} & \textbf{Artifacts} & \textbf{Metric} \\
        \midrule
        TheoremQA     & Skill        & 0     & 747   & 1    & Accuracy \\
        LogicBench    & Skill        & 0     & 760   & 1    & Accuracy \\
        MedCalc-Bench & Skill        & 0     & 1,100 & 1    & Accuracy \\
        CHAMP         & Skill        & 0     & 223   & 1--5 & Accuracy \\
        BigCodeBench  & Skill        & 4,000 & 667   & 2--5 & pass@1 \\
        \midrule
        MuSiQue       & GT document  & 4,000 & 2,417 & 2--4 & F1 / EM \\
        \midrule
        PersonaMem-v2 & GT message   & 4,000 & 5,000 & 2--8 & Accuracy \\
        \bottomrule
    \end{NiceTabular}
\end{table}

\paragraph{Skills.}
We use the task-annotated skills and prompt builders from SRA-Bench~\citep{su2026skill} for TheoremQA~\citep{chen2023theoremqa}, LogicBench~\citep{parmar2024logicbench}, MedCalc-Bench~\citep{khandekar2024medcalc}, CHAMP~\citep{mao2024champ}, and BigCodeBench~\citep{zhuo2025bigcodebench}. Each annotated skill is treated as a separate artifact. TheoremQA, LogicBench, and MedCalc-Bench contain one skill per request. CHAMP contains a mixture: 114 requests have one skill, and 109 have multiple skills. It should therefore not be interpreted as an exclusively multi-artifact benchmark. ToolQA is excluded because its single skill is reused across multi-step ReAct interactions, which differs from the request-level setting studied here.

For BigCodeBench, we use a fixed skill-disjoint partition of the 1,140 SRA-Bench instances. The manifest divides 139 skill IDs into 70 training-side and 69 evaluation-side skills. An instance is retained only if all its annotated skills belong to the same side. This yields 198 real training-side instances and 667 evaluation instances; 275 instances spanning both partitions are excluded. The main adapter mixture uses synthetic tasks built from the training-side skills, rather than the 198 real instances. The latter support separate validation and ablation experiments.

\paragraph{Documents.}
For MuSiQue~\citep{trivedi2022musique}, we retain only paragraphs marked as supporting the answer, preserving their original relative order and document indices. Each supporting document is one artifact. The 2,417 evaluation requests contain 1,252 two-document, 760 three-document, and 405 four-document contexts. Questions, reference answers, and answer aliases are retained from the official data. This GT-document protocol isolates the use of supplied evidence; it is distinct from evaluation with the full candidate-document pool.

\paragraph{Memory messages.}
For PersonaMem-v2~\citep{jiang2025personamemv2}, we use the annotated \texttt{related\_conversation\_snippet} rather than the full conversation history. Each annotated user or assistant message is a separate artifact, with its role, content, and annotation order preserved. The request-specific question and multiple-choice instruction remain online. We retain the saved per-request option ordering and answer mapping for all methods. The evaluation uses all 5,000 benchmark requests; unannotated conversation history and persona background are not added.

\subsection{Evaluation Metrics and Efficiency Protocol}
\label{app:metrics}

\paragraph{Task quality.}
We use the SR-Agents benchmark scoring pipeline for the five skill tasks. BigCodeBench pass@1 is the fraction of requests whose single generated solution passes the task's unit tests, executed in local subprocesses. For MuSiQue, the final answer is extracted from the last \texttt{ANSWER:} line, or the last nonempty line when that marker is absent, and scored against the reference answer and aliases using the benchmark's answer metrics. F1 is the main-table metric; EM is also reported for diagnostic experiments. PersonaMem-v2 uses the official multiple-choice answer extraction and correctness functions with the saved option mapping. A response that reaches its generation limit is scored as generated rather than discarded.

\Needspace{22\baselineskip}
\begin{wraptable}{r}{0.50\linewidth}
    \centering
    \caption{\textbf{Warm-cache TTFT boundary.} Reusable KV is already resident in CPU memory before timing begins.}
    \label{tab:app-ttft-boundary}
    \begingroup
    \footnotesize
    \setlength{\tabcolsep}{4pt}
    \renewcommand{\arraystretch}{1.15}
    \begin{NiceTabular}{X[l]c}[width=\linewidth]
\CodeBefore
    \rowcolor{black!6}{1}
\Body
        \toprule
        \textbf{Operation} & \textbf{Timed?} \\
        \midrule
        Model loading, tokenization, artifact-boundary identification & No \\
        Offline artifact KV construction & No \\
        CPU RAM cache lookup and GPU transfer & Yes \\
        KV assembly and online computation, including repair & Yes \\
        Greedy first-token selection & Yes \\
        Subsequent generation, text decoding, and scoring & No \\
        \bottomrule
    \end{NiceTabular}
    \endgroup
\end{wraptable}

\paragraph{Timing boundary.}
We measure warm-cache time to first token (TTFT) with reusable KV resident in CPU RAM. Model loading, tokenization, artifact-boundary identification, and cache construction occur before the timed region. Timing begins before cache lookup and includes transfer to the GPU, KV assembly, method-specific online repair, and uncached-token prefill. It ends after greedy selection of the first output token. GPU operations are synchronized at both boundaries. Subsequent generation, text decoding, and scoring are excluded (Table~\ref{tab:app-ttft-boundary}). Thus, this metric measures request processing from a prepared tokenized input, rather than end-to-end service latency.

\paragraph{Measurement and aggregation.} For the main-table efficiency results, we evaluate the full evaluation set of each dataset with batch size one and no concurrent workload on the same GPU. The dedicated efficiency runs use batch size one and generate exactly one token per request. Each worker performs five untimed warm-up requests for each dataset--method pair, followed by one timed measurement per evaluation request. Quality generation is evaluated separately; concurrent quality-run timings are not used as formal TTFT results. For dataset $d$ and method $m$, we report
\begin{equation}
T_{m,d}=\operatorname{median}_{i\in\mathcal{E}_d} t_{m,d,i},
\qquad
S_{m,d}=\frac{T_{\mathrm{Full},d}}{T_{m,d}},
\label{eq:app-ttft-aggregation}
\end{equation}
where $\mathcal{E}_d$ is the matched evaluation cohort and $t_{m,d,i}$ is the measured TTFT for request $i$. When requests are distributed across workers, we merge request-level timings before computing the median; we do not average worker medians. Cross-dataset efficiency summaries use the geometric mean over the same datasets for every method: $G_m=(\prod_{d\in\mathcal{D}}T_{m,d})^{1/|\mathcal{D}|}$, with aggregate speedup $G_{\mathrm{Full}}/G_m$.

\paragraph{Cache and execution settings.} A token-identical prefix shared across the dataset may also be cached before timing. For \method, query updates are merged into projection weights for online inference, while artifact KV is built with the unadapted model. Full and cached methods must use matched prompts, evaluation cohorts, hardware, and timing boundaries for a valid speedup comparison.

\subsection{Prompt Construction and Artifact Boundaries}
\label{app:prompts}

We render requests with the corresponding Qwen3 chat template and \texttt{enable\_thinking=False}. Artifact boundaries are located within the rendered token sequence. Full and PIC receive identical prompt tokens within each comparison; they differ in how the artifact states are computed. Text outside the artifact spans, including instructions, separators, and questions, is processed online.

\begin{table}[!htbp]
    \centering
    \small
    \renewcommand{\arraystretch}{1.15}
    \caption{\textbf{Prompt layouts across artifact families.}
    Bracketed artifact spans are cached; other text is computed online. These layouts omit full task text and chat-template special tokens.}
    \label{tab:app-prompt-layouts}
    \vspace{2pt}
    \begin{NiceTabular}{lX[l]}[width=\linewidth]
\CodeBefore
    \rowcolor{black!6}{1}
\Body
        \toprule
        \textbf{Family} & \textbf{Model-visible layout} \\
        \midrule
        \textbf{Skills} & Optional system instruction; \texttt{Relevant Skill:};
        [skill A]; separator; [skill B]; task and answer instruction. \\
        \addlinespace[3pt]
        \textbf{Documents} & System instruction; \texttt{Documents:};
        [document A: index, title, text]; [document B: index, title, text];
        question and answer instruction. \\
        \addlinespace[3pt]
        \textbf{Memories} & [relevant message A]; [relevant message B];
        user query; multiple-choice instruction with fixed options. \\
        \bottomrule
    \end{NiceTabular}
\end{table}

\paragraph{Skill and document spans.}
For SRA-Bench, the skill contents are prepended to the task using the benchmark prompt builder. The standard layout separates consecutive skills with \texttt{\textbackslash n---\textbackslash n}; these separators are outside the cached skill spans. MuSiQue artifacts contain the document identifier, title, and text. The surrounding task instructions and question are not part of the cached documents. Any separately reported separator-free efficiency sweep uses an explicitly different prompt protocol and should not be combined with quality scores from the standard layout.

\paragraph{Message spans.} PersonaMem-v2 artifacts cover complete historical messages, including their chat-template boundaries. This preserves user/assistant roles when independently encoded blocks are assembled. Only annotated messages are cached; the new question and option instruction are not.

\paragraph{Cache construction and reuse.}
Each artifact is independently encoded with query adapters disabled. At request time, its cached KV is assembled in the original prompt order during evaluation. Adapters are enabled for uncached prompt tokens and generated tokens, while stored artifact KV remains fixed. Subsequent online states and their newly produced KV evolve under the adapted computation.

\subsection{Training Data Construction}
\label{app:training-data}

\paragraph{Balanced mixture.}
We form one fixed mixture of 12,000 requests: 4,000 MuSiQue questions, 4,000 synthetic BigCodeBench-style tasks, and 4,000 PersonaMem-v2 questions. The two model sizes use the same selected request IDs but their own Full-context teacher responses. MuSiQue requests are selected from an 8,192-example pool drawn from the official training split. PersonaMem-v2 training personas exclude all evaluation personas; validation personas are additionally removed before sampling the training mixture. TheoremQA, LogicBench, MedCalc-Bench, and CHAMP contribute no training examples to this mixture.

We reserve 128 MuSiQue training-split requests, 31 real training-side BigCodeBench requests, and 128 PersonaMem-v2 requests for validation. The selected training and validation request IDs are disjoint. For PersonaMem-v2, training, validation, and evaluation are also disjoint at the persona level. These main-training validation sets should not be conflated with separately constructed diagnostic cohorts.

\paragraph{Synthetic skill-use tasks.}
We construct the BigCodeBench-style source using only training-side skills. The synthesis procedure samples two to five skill documents and asks a generator to produce a Python function task requiring their joint use. The task specifies its behavior, arguments, return type, and edge cases, and includes a starter code block defining \texttt{task\_func}. The parser requires a valid starter, retains imports and the function signature, and replaces the function body with \texttt{pass} so that a completed solution is not embedded in the input. The resulting tasks are reformatted with the same skill prompt builder used for evaluation. All selected skill IDs belong to the training partition; evaluation-side skill documents are excluded. The 4,000 synthetic requests are shared across model sizes. They provide distillation inputs rather than independently verified programming solutions: supervision comes from each model's Full teacher.

The recovered generation pipeline uses Qwen3-4B with task-generation temperature $0.8$, top-$p=0.95$, and a 512-token task budget. The number of skills is sampled from $\{2,3,4,5\}$ with probabilities $\{0.30,0.35,0.25,0.10\}$. Tasks with an invalid starter or prompts exceeding 36,864 tokens are rejected. The source pipeline also requires a nonempty Full response containing a code block, using a 1,024-token greedy response budget; it does not validate synthetic solutions with unit tests. The completed source combines 3,027 initially accepted tasks with 973 additional tasks generated in three shards. 

\paragraph{Teacher supervision.}
For each model, a frozen copy of the base model produces greedy Full-context responses with thinking disabled. Training teacher-forces the saved response tokens in both branches and minimizes teacher-to-student output-distribution KL on response positions. Prompt positions are not directly supervised. The token losses are averaged within each response and then across requests, giving each request equal weight rather than weighting it by response length. The student uses independently cached artifact KV.

The bundled teacher-response pool has maximum retained prefix lengths of 1,024 tokens for BigCodeBench and 256 for PersonaMem-v2. MuSiQue prefixes have limits of 128 tokens for Qwen3-4B and 256 for Qwen3-8B. A retained prefix need not be a complete answer. In particular, all 4,000 selected Qwen3-4B MuSiQue trajectories are legacy fixed prefixes; 3,793/4,000 Qwen3-4B and 3,907/4,000 Qwen3-8B PersonaMem-v2 trajectories reach their 256-token limit. These training-prefix limits are separate from evaluation generation limits.

\subsection{Training and Generation Settings}
\label{app:implementation}

\Needspace{22\baselineskip}
\begin{wraptable}{r}{0.50\linewidth}
    \centering
    \caption{\textbf{Shared training configuration for Qwen3-4B and Qwen3-8B.} Global batch size includes all workers and accumulation steps.}
    \label{tab:app-training-settings}
    \begingroup
    \footnotesize
    \setlength{\tabcolsep}{4pt}
    \renewcommand{\arraystretch}{1.12}
    \begin{NiceTabular}{lX[l]}[width=\linewidth]
\CodeBefore
    \rowcolor{black!6}{1}
    \rowcolor{black!6}{2,6,11}
\Body
        \toprule
        \textbf{Setting} & \textbf{Value} \\
        \midrule
        \multicolumn{2}{l}{\textit{Adapter}} \\
        Projection & Query only \\
        Layers & L0--35 (all layers) \\
        Rank / scale & $r=8$, $\alpha=16$ \\
        \multicolumn{2}{l}{\textit{Optimization}} \\
        Optimizer & AdamW \\
        Learning rate & $2\times10^{-4}$ \\
        Weight decay & $0.01$ \\
        Gradient clipping & Max. norm $1.0$ \\
        \multicolumn{2}{l}{\textit{Training schedule}} \\
        Global batch & 16 requests \\
        Micro-batch / GPU & 1 request \\
        Duration & 1 epoch / 750 updates \\
        Checkpoint & Final epoch \\
        Mixture & 4,000 per source \\
        Precision & BF16 \\
        \bottomrule
    \end{NiceTabular}
    \endgroup
\end{wraptable}

We use Hugging Face Transformers and BF16 computation. Query adapters target \texttt{q\_proj} in every Transformer layer. The low-rank matrix $A_{\ell}$ is initialized with Kaiming uniform initialization and $B_{\ell}$ with zeros, so the initial update is zero. All base-model parameters are frozen. Table~\ref{tab:app-training-settings} summarizes the shared training settings for both model sizes.

\paragraph{Generation.}
Evaluation uses greedy decoding with thinking disabled and one request per model forward. Independent workers can process different requests in parallel for quality evaluation. All methods use the same dataset-specific generation budget and retain the benchmark's required answer format. Quality-run timings are not used as formal latency measurements.

\paragraph{Deployment.}
The learned update can be merged into the query projection as $W_{\ell}^{Q} \leftarrow W_{\ell}^{Q} + (\alpha/r)B_{\ell}A_{\ell}$ for online inference. Artifact caches are constructed with the original, unadapted model. The main PIC and \method runtime uses PyTorch scaled dot-product attention (SDPA), with causal masks and position IDs specified for the assembled context.

\subsection{Analysis Settings}
\label{app:analysis-settings}

\paragraph{Distractor evaluation.}
The quality comparison in Figure~\ref{fig:pic-observations}(a) uses fixed cohorts of 667 BigCodeBench examples and 2,401 MuSiQue examples. All annotated relevant artifacts are retained, and distractor sets are nested across conditions. For BigCodeBench, we increase the target total number of skills to 4, 8, 12, and 16, appending non-annotated skills from the evaluation-side skill pool. At the smallest setting, nine examples retain five skills because all five are annotated as relevant. For MuSiQue, we use examples with 20 candidate documents and add 0, 2, 4, 8, or 16 non-supporting documents. Full and PIC receive identical inputs within each condition.

\paragraph{Attention-output alignment.}
The analysis in Figure~\ref{fig:training-output-alignment} uses 128 examples each from MuSiQue and BigCodeBench. Full, PIC, and \method receive identical prompts and condition on the same saved Full-response prefixes. We compare per-head attention outputs before the output projection, separately for online prompt and answer positions. Similarities are averaged over positions within each example, then over heads and examples; overall summaries additionally average over layers. Each example receives equal weight.
\clearpage
\section{Routing Recovery Case Studies}
\label{app:routing-analysis}
\label{app:routing-case-studies}

We extend the main-text MuSiQue analysis with one example each from CHAMP, BigCodeBench, and PersonaMem-v2. The BigCodeBench and PersonaMem-v2 examples were selected before inspecting their attention tensors.

\Needspace{22\baselineskip}
\begin{wraptable}{r}{0.44\linewidth}
    \centering
    \caption{\textbf{Attention-output recovery.} Similarity to Full at the final prompt token, averaged over layer--head pairs. Each row is one example.}
    \label{tab:routing-case-recovery}
    \begingroup
    \footnotesize
    \setlength{\tabcolsep}{2pt}
    \renewcommand{\arraystretch}{1.12}
    \begin{NiceTabular}{X[l]rrrr}[width=\linewidth]
\CodeBefore
    \rowcolor{black!6}{2}
\Body
        \toprule
        & \multicolumn{2}{c}{\textbf{Cosine} $\uparrow$}
        & \multicolumn{2}{c}{\textbf{Rel.\ L2} $\downarrow$} \\
        \cmidrule(lr){2-3}\cmidrule(lr){4-5}
        \textbf{Dataset} & \textbf{PIC} & \textbf{Oracle}
        & \textbf{PIC} & \textbf{Oracle} \\
        \midrule
        CHAMP         & .859 & .988 & .531 & .131 \\
        BigCodeBench  & .892 & .989 & .474 & .124 \\
        PersonaMem-v2 & .910 & .985 & .418 & .140 \\
        \bottomrule
    \end{NiceTabular}
    \endgroup
\end{wraptable}

\begin{figure}[h]
    \centering
    \includegraphics[width=0.88\linewidth]
    {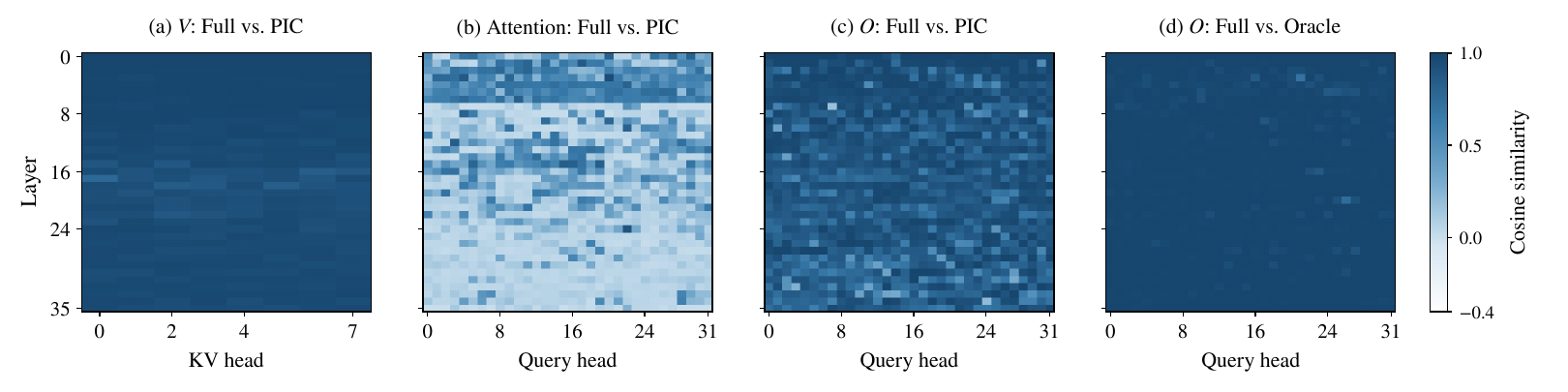}

    \par\vspace{2pt}
    \includegraphics[width=0.88\linewidth]
    {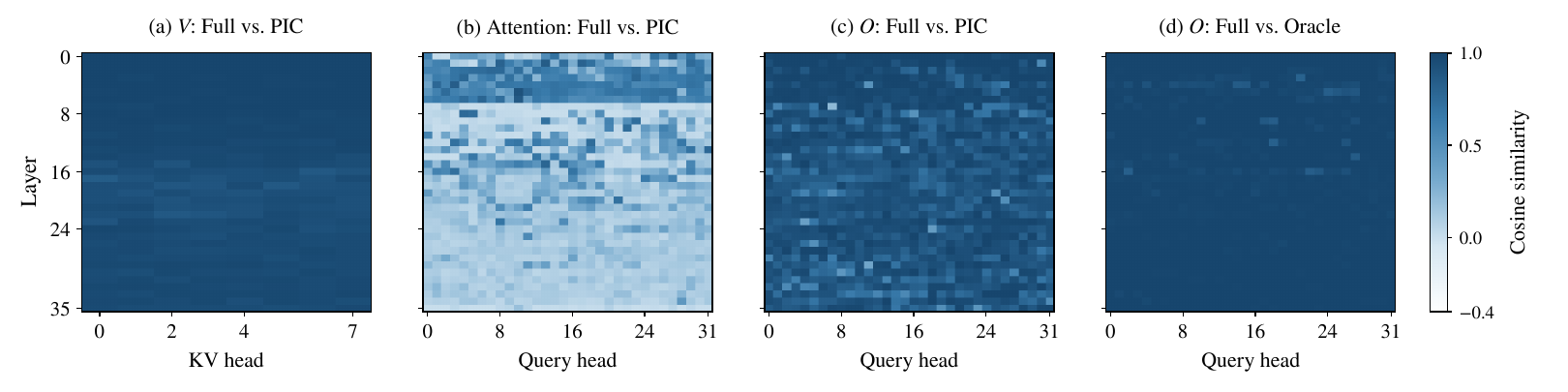}

    \par\vspace{2pt}
    \includegraphics[width=0.88\linewidth]
    {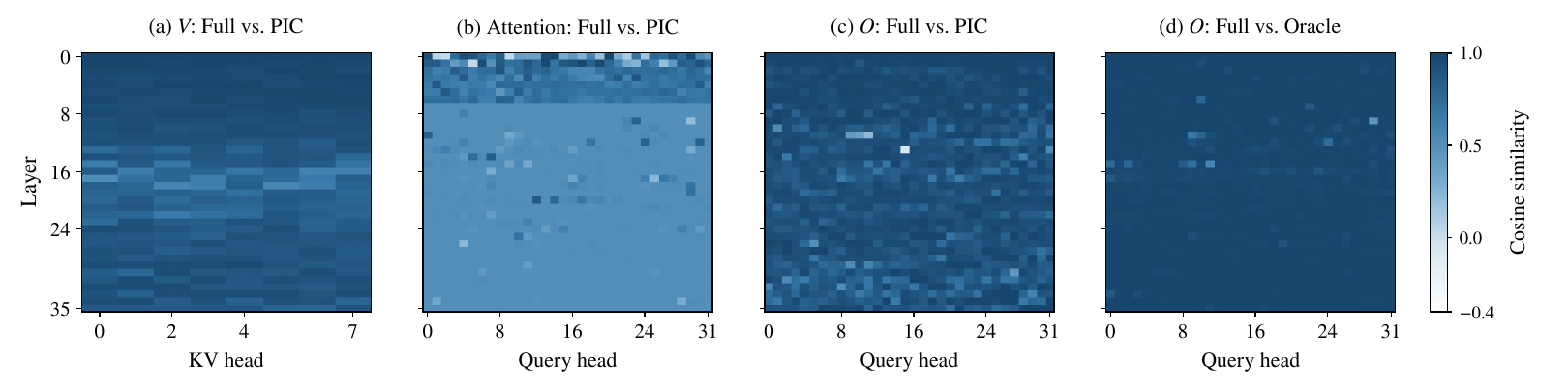}

    \caption{\textbf{Routing recovery across artifact types.} Rows: CHAMP, BigCodeBench, and PersonaMem-v2. Columns: Full--PIC cosine similarity for (a) cached values, (b) artifact attention, and (c) outputs; (d) Full--oracle output similarity. Panels (b--d) use the final prompt token. All panels share the same color scale.}
    \label{fig:routing-additional-cases}
\end{figure}

\paragraph{Measurements.}
Figure~\ref{fig:routing-additional-cases} compares Full and PIC for cached values, attention over artifact tokens, and attention outputs; the final panel compares oracle outputs with Full. Value cosine similarity is computed for each layer and KV head by concatenating value vectors across artifact tokens. Attention and output similarities are computed for each layer and query head at the final prompt token, with outputs measured before the output projection. All panels share a cosine scale of $[-0.4,1]$.

\paragraph{Recovery with unchanged artifact KV.}
The oracle replaces complete attention rows with the Full reference while preserving artifact KV; online states and their values evolve under the intervention. Across all three examples, this raises mean output cosine similarity to at least $0.985$ and reduces relative L2 error (Table~\ref{tab:routing-case-recovery}). These cases illustrate that independently cached values can support outputs close to Full when read with restored routing. They do not establish that value changes are harmless or that exact Full representations are recovered. The visualizations concern one query position per example; task-level recovery is evaluated separately in the main text.

\clearpage
\section{Baseline Implementations}
\label{app:baselines}

We evaluate all methods with the Qwen3 backbones, prepared evaluation
requests, and scoring procedures described in Appendix~\ref{app:details}. Table~\ref{tab:app-baseline-overview} summarizes their mechanisms and configurations. We adapt the available public implementations to Qwen3; AgentKVShift is a paper-based reimplementation because we could not identify an author-released repository. The reported comparisons concern these implementations under our common evaluation protocol, rather than the original systems' serving performance.

\begin{table}[!htbp]
    \centering
    \caption{\textbf{Baseline mechanisms and configurations.}
    Layer indices are zero-based. COMB corresponds to the
    \texttt{encoder\_pic} implementation name.}
    \label{tab:app-baseline-overview}
    \small
    \setlength{\tabcolsep}{5pt}
    \renewcommand{\arraystretch}{1.16}
    \begin{NiceTabular}{lX[l]X[l]}[width=\linewidth]
\CodeBefore
    \rowcolor{black!6}{1}
    \rowcolor{black!6}{2,7}
\Body
        \toprule
        \textbf{Method} & \textbf{Mechanism} & \textbf{Configuration} \\
        \midrule
        \multicolumn{3}{l}{\textit{Training-free}} \\
        APE & Calibrated attention over independently encoded blocks
        & Temperature $0.9$; scale $0.9$ \\
        EPIC & Recompute artifact boundary tokens
        & First 4 tokens per eligible artifact \\
        CacheBlend & Selective recomputation based on value deviation
        & Ratio $0.16$; check layer L1; original-position causal mask \\
        AgentKVShift & Probe-based key/value correction
        & Probe ratio $0.10$; check layer L1 \\
        \midrule
        \multicolumn{3}{l}{\textit{Training-based}} \\
        KVLink & Learned link tokens and model fine-tuning
        & 5 link tokens per artifact; response CE \\
        COMB & Independent encoder and gated cross-attention
        & 9 encoder layers and cross-attention insertion points; response CE \\
        \bottomrule
    \end{NiceTabular}
\end{table}

\subsection{Reference Methods}
\label{app:baseline-controls}

\paragraph{Full.}
Full causally prefills the complete request, allowing each artifact to condition on its actual preceding context. It provides the reference
quality and latency for each matched evaluation cohort.

\paragraph{PIC.}
PIC independently encodes each artifact using local positions starting at zero. At inference, cached blocks are assembled in prompt order, while uncached text is computed with its causal history. Artifact tokens are not recomputed. The efficiency implementation processes online positions layer by layer, as described in Appendix~\ref{app:metrics}.

\paragraph{PIC-Pos.}
This diagnostic control independently encodes artifacts at their target positions in the assembled prompt. It changes positional coordinates without restoring cross-artifact conditioning. Because cache construction requires the target placement, PIC-Pos is a position control rather than a placement-independent cache-building strategy.

\subsection{Training-Free Baselines}
\label{app:training-free-baselines}

\paragraph{APE.}
APE~\citep{yang2025ape} separately computes attention over cached context and prefix/online tokens, then combines their outputs using calibrated log-sum-exp terms. We use temperature $0.9$ and scale $0.9$, following the public implementation.\footnote{\url{https://github.com/Infini-AI-Lab/APE}} Our port supports Qwen3 query/key normalization and RoPE, using fused attention outputs and log-sum-exp values where supported. Group normalization is computed in FP32. A zero-length shared prefix is allowed for prompts that begin directly with cached artifacts.

Cache boundaries matter for this comparison. The original port can include inter-artifact separators in independently encoded context blocks. In the BigCodeBench and TheoremQA boundary-controlled runs, only artifact content is cached: separators are processed online with their causal history. We retain APE's shared-prefix encoding and attention calibration. These runs are a cache-boundary adaptation of APE, and their quality and TTFT must be paired within that layout. APE calibrates attention during online reading; it does not first run PIC and then recompute artifact tokens.

\paragraph{EPIC.}
Following EPIC's public LegoLink prototype,\footnote{\url{https://github.com/DerekHJH/epic}} we recompute the first four tokens of each eligible artifact from the first Transformer layer~\citep{hu2024epic}. An artifact beginning at position zero is skipped because it has no preceding context to restore. All online prompt tokens are also computed. Cached keys are stored before RoPE and rotated to their target positions during assembly; sparse attention masks follow each query's original position. Four tokens is one supported configuration, rather than a universal EPIC default. The prototype's internal name \texttt{kvlink-4} denotes this boundary repair and is unrelated to the trained KVLink baseline below.

\paragraph{CacheBlend.}
CacheBlend~\citep{yao2024cacheblend} ranks token positions by the difference between freshly computed and cached values, then propagates selected tokens through subsequent layers. We use check layer L1 and a selection ratio of $0.16$, retaining the complete online suffix. At the check layer, selection occurs before attention; later layers update the selected positions while reusing the remaining cached states.

Our final Qwen3 variant uses a causal mask indexed by the queries' original token positions. The inspected public prototype instead uses a bottom-right-aligned mask for sparse queries; the two masks need not permit the same keys.\footnote{\url{https://github.com/YaoJiayi/CacheBlend}} We also preserve the entire suffix, correcting the prototype example's one-token suffix-length calculation. We therefore report this method as a Qwen3 adaptation of CacheBlend. Results from the earlier mask variant are not interchangeable with the revised variant's quality or timing.

\paragraph{AgentKVShift.}
We implement the paper's probe-based correction~\citep{agentkvshift}. Within each artifact, key and value probes are selected separately using deviation scores, with a probe ratio of $0.10$. Their union determines which tokens are recomputed. Layer-wise probe residuals estimate key and value corrections for the remaining cached tokens; key and value writebacks retain their respective probe sets.

The Qwen3 port concatenates KV heads when computing per-token deviations and compares keys in aligned RoPE coordinates. We use check layer L1, the first KV projection whose input has passed through a preceding attention layer. Since this extraction convention is not established by author-released code, it is an explicit implementation choice. Attention visibility follows original token positions.

\subsection{Training-Based Baselines}
\label{app:training-based-baselines}

\paragraph{KVLink.}
KVLink~\citep{kvlink} introduces five learned link tokens per artifact and fine-tunes the backbone.\footnote{\url{https://github.com/UCSB-NLP-Chang/KVLink}} Its block attention mask keeps ordinary artifact tokens within their own causal block, while link and online tokens can attend to preceding valid blocks. We extend the Qwen3 tokenizer with link tokens and train all model weights, including the new embeddings. At inference, artifact states are cached using the trained model and link tokens are computed online. This requires a method-specific cache; it does not retain the unadapted base model's artifact KV.

\paragraph{COMB.}
COMB~\citep{encoderpic} independently encodes artifacts and exposes their representations through added cross-attention modules.\footnote{\url{https://github.com/shijuzhao/Comb}} Our Qwen3 port uses one bidirectional encoder layer and one cross-attention block per four original decoder layers. For both 36-layer backbones, cross-attention is inserted before L3, L7, $\ldots$, L35, giving nine insertion points. The original backbone and embeddings remain frozen; the new encoder, KV projections, cross-attention, MLPs, and gates are trained. Cross-attention and MLP gates are initialized to zero. The cached states are encoder-produced representations, rather than ordinary decoder artifact KV. We use the name COMB throughout the paper;
\texttt{encoder\_pic} is only the implementation identifier.

\Needspace{22\baselineskip}
\begin{wraptable}{r}{0.48\linewidth}
    \centering
    \caption{\textbf{Training settings for KVLink and COMB.}
    Both models use the same 12K requests and model-specific teacher
    responses as \method.}
    \label{tab:app-baseline-training}
    \begingroup
    \footnotesize
    \setlength{\tabcolsep}{4pt}
    \renewcommand{\arraystretch}{1.12}
    \begin{NiceTabular}{X[l]cc}[width=\linewidth]
\CodeBefore
    \rowcolor{black!6}{1}
\Body
        \toprule
        \textbf{Setting} & \textbf{KVLink} & \textbf{COMB} \\
        \midrule
        Learning rate & $5\times10^{-6}$ & $5\times10^{-5}$ \\
        Optimizer & \multicolumn{2}{c}{AdamW} \\
        Weight decay & \multicolumn{2}{c}{$0.01$} \\
        Gradient norm cap & \multicolumn{2}{c}{$1.0$} \\
        Global batch & \multicolumn{2}{c}{16 requests} \\
        Micro-batch / GPU & \multicolumn{2}{c}{1 request} \\
        Duration & \multicolumn{2}{c}{1 epoch / 750 updates} \\
        Precision & \multicolumn{2}{c}{BF16} \\
        \bottomrule
    \end{NiceTabular}
    \endgroup
\end{wraptable}

\paragraph{Matched data and training budget.}
Both baselines use the same 12,000 training requests and corresponding Full-context teacher responses as \method. They minimize response-token cross-entropy, retaining their supervised language-modeling objectives rather than adopting \method's distributional KL loss. Prompt tokens are excluded from the loss. Each model size receives a separately trained checkpoint.

We train for one epoch with a global batch of 16 requests and one request per GPU forward, yielding 750 optimizer updates. AdamW uses constant learning rates of $5\times10^{-6}$ for KVLink and $5\times10^{-5}$ for COMB, weight decay $0.01$, and gradient clipping at norm $1.0$ (Table~\ref{tab:app-baseline-training}). We evaluate the final checkpoint. The shared data and training duration are our controlled comparison protocol, not a reproduction of the original papers' complete training corpora or schedules. The methods also differ in trainable parameter count, so equal numbers of examples and updates do not imply equal training compute.

\paragraph{Efficiency comparisons.}
Each method constructs its own reusable states before timing and places them in CPU RAM. Online transfer, cache assembly, link-token computation, cross-attention, and any repair are included when applicable, using the TTFT boundary in Appendix~\ref{app:metrics}. Offline construction and training are excluded. CacheBlend and APE quality and efficiency must use the same mask and cache-layout variants, respectively.

\clearpage
\section{Supplementary Analyses}
\label{app:supplementary-analyses}

\subsection{Alternative Repair Interventions}
\label{app:alternative-repairs}

We first examine whether positional alignment restores effective cache use, then compare component-level interventions under a matched protocol. These Qwen3-4B diagnostics complement the task-level routing oracle in the main text.

\begin{figure}[h]
    \centering
    \includegraphics[width=0.88\linewidth]
    {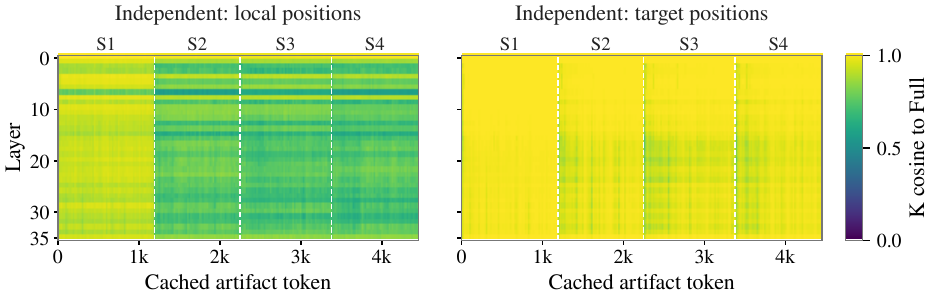}
    \caption{
    \textbf{Position alignment leaves a residual quality gap.} Key cosine similarity to Full for \texttt{champ\_00004}, using local positions (left) or target positions (right). S1--S4 mark skill boundaries. Both panels average all KV heads and share a color scale.
    }
    \label{fig:alternative-position-alignment}
\end{figure}

\paragraph{Position alignment improves keys without guaranteeing recovery.}
For the four-skill CHAMP example \texttt{champ\_00004}, PIC-Pos increases mean key cosine similarity to Full from 0.794 to 0.975 by independently encoding artifacts at their target prompt positions (Figure~\ref{fig:alternative-position-alignment}). Nevertheless, both PIC and PIC-Pos answer incorrectly, whereas Full answers correctly. Thus, closer key representations do not guarantee task recovery in this example: position alignment does not restore the contextual conditioning omitted by independent encoding.

The heatmaps average all eight KV heads and group at most 16 adjacent tokens within each artifact for display. PIC-Pos re-encodes artifacts at target positions rather than rotating an existing cache; its values are not assumed to be bitwise identical to those from local-position encoding.

\FloatBarrier
\Needspace{23\baselineskip}
\begin{wraptable}{r}{0.48\textwidth}
    \centering
    \small
    \caption{
    \textbf{All-layer component interventions.} Mean output KL to Full; lower is better. Full KV and Full $O$ are separate reconstruction controls.
    }
    \label{tab:component-interventions}
    \setlength{\tabcolsep}{4pt}
    \renewcommand{\arraystretch}{1.08}
    \begin{NiceTabular}{@{}X[l]rr@{}}[width=\linewidth]
        \toprule
        Intervention & MuSiQue & BCB \\
        \midrule
        PIC          & 0.8480  & 0.1221 \\
        PIC-Pos      & 0.8536  & 0.0850 \\
        Full $K$     & 0.8104  & 0.0447 \\
        Full $V$     & 13.5594 & 8.4155 \\
        Full routing & \textbf{0.1059} & \textbf{0.0086} \\
        Full KV / Full $O$ & 0.0000 & 0.0000 \\
        \bottomrule
    \end{NiceTabular}
\end{wraptable}

\paragraph{Matched interventions isolate alternative recovery paths.}
We evaluate 32 fixed randomly sampled examples each from MuSiQue and BigCodeBench. All branches use identical prompts and the same saved Full-response prefixes, capped at 64 tokens. Starting from PIC, we separately replace artifact $K$, $V$, or both with their Full counterparts before online prefill. We also restore Full attention rows (routing) or complete attention outputs ($O$, before the output projection) at online prompt positions after the first artifact and subsequent response inputs. All interventions span L0--35. Routing restoration retains artifact KV and applies Full attention weights to the intervention branch's own values; online states evolve under the intervention.

We measure $D_{\mathrm{KL}}(p^{\mathrm{Full}} \Vert p^{\mathrm{intervention}})$, averaging over response positions within each request and then equally across requests. This measures output-distribution recovery under shared prefixes, rather than free-generation task accuracy.

\begin{figure}[h]
    \centering
    \includegraphics[width=0.94\linewidth]
    {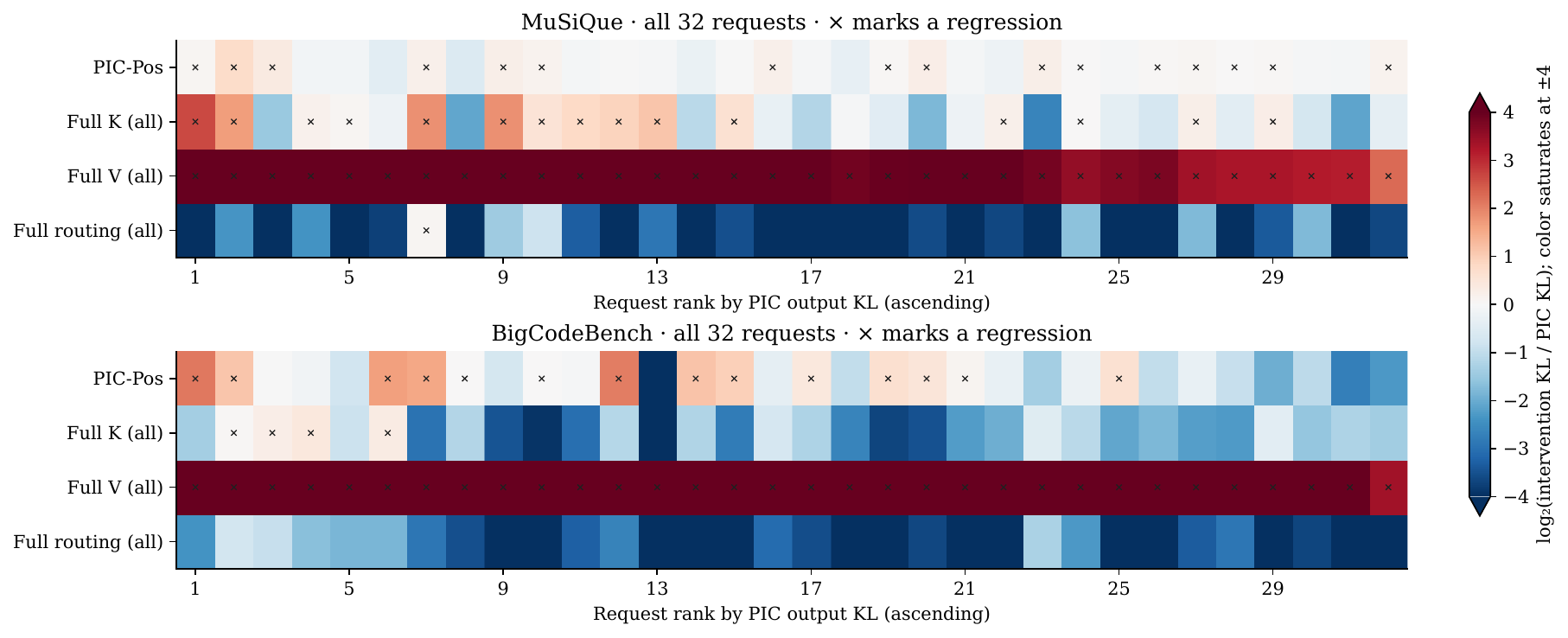}
    \caption{
    \textbf{Recovery across individual requests.} Columns represent examples ordered by PIC output KL within each dataset. Colors show $\log_2(D_{\mathrm{KL}}^{\mathrm{intervention}}/ D_{\mathrm{KL}}^{\mathrm{PIC}})$: blue indicates improvement and red deterioration. Crosses mark regressions; colors saturate at $\pm4$. All component interventions span every layer.
    }
    \label{fig:component-sample-recovery}
\end{figure}

\Needspace{10\baselineskip}
\paragraph{Restoring routing consistently reduces output divergence.} Full routing reduces mean output KL by \textbf{87.5\% on MuSiQue and 93.0\% on BigCodeBench} (Table~\ref{tab:component-interventions}). Paired bootstrap 95\% intervals are [81.4\%, 92.8\%] and [88.0\%, 95.8\%], respectively, using 2,000 request-level resamples. The reduction occurs in 31/32 MuSiQue examples and all 32 BigCodeBench examples (Figure~\ref{fig:component-sample-recovery}). In contrast, replacing only $K$ produces dataset-dependent gains, while replacing only $V$ substantially increases KL. Full KV and Full $O$ recover Full logits to numerical precision, serving as positive reconstruction controls.

\FloatBarrier
\paragraph{Implications for cache-preserving recovery.}
Together, these results show that \textbf{restoring routing can substantially recover Full-like output distributions without replacing cached artifact values}. They do not establish that values are unimportant: the poor V-only result instead shows that inserting Full values into the remaining PIC computation can be disruptive. Nor does improved representation similarity guarantee behavioral recovery: under all-layer $K$ replacement, 14/32 MuSiQue examples exhibit higher mean answer-position $O$ cosine similarity but worse output KL. These diagnostics therefore support adapting how the cache is read, while the main-text routing oracle provides the corresponding task-level evidence.
\subsection{Limitations and Future Directions}
\label{app:limitations}

Our evaluation covers skills, documents, and memories as separate task settings, but does not examine requests that combine all three artifact types. We focus on PIC in models with full attention; applicability to other attention architectures remains unexplored. Our contribution concerns effective cache reading rather than cache storage or transport. Although our TTFT measurements include CPU-to-GPU cache transfer, we do not optimize cache placement, eviction, or communication. Looking ahead, our results motivate exploring cache reading as a post-training objective, enabling models to better use context prepared independently of the current request. Combining \method with KV compression and retrieval-based artifact selection is a promising direction for extending cache reuse to larger contexts and tighter memory budgets.
\clearpage
\section{Detailed Results by Evaluation Domain}
\label{app:domain-results}

Tables~\ref{tab:app-indomain-results} and~\ref{tab:app-ood-results} reproduce the in-domain and out-of-domain portions of Table~\ref{tab:multi-artifact-results} separately for readability. All scores, timings, aggregate values, and rankings are unchanged. In-domain tasks are BigCodeBench (BCB), MuSiQue (MSQ), and PersonaMem-v2 (PMem); out-of-domain tasks are TheoremQA (TQA) and CHAMP (CHP), neither of which is included in adapter training. The domain distinction concerns training coverage, not artifact count: TheoremQA uses one artifact, while CHAMP contains both single- and multiple-artifact requests.

\begin{table}[!htbp]
\centering
\caption{\textbf{In-domain quality and efficiency.}
Scores are pass@1 for BCB, F1 for MSQ, and accuracy for PMem (all in percent). TTFT is median warm-cache latency in milliseconds. Avg is the arithmetic mean of scores; $\Delta$ is its change from Full in percentage points. Geo. is the geometric mean of TTFT, and speedup is relative to Full. Full and PIC are excluded from ranking; bold and underlined entries mark the best and second-best results among the other methods, including ties.}
\label{tab:app-indomain-results}
\begingroup
\small
\setlength{\tabcolsep}{3pt}
\renewcommand{\arraystretch}{1.10}
\begin{adjustbox}{max width=\linewidth}
\begin{NiceTabular}{@{}l*{10}{r}@{}}
\CodeBefore
    \rowcolor{black!6}{3,15}
    \rowcolor{black!6}{6,11,18,23}
    \rowcolor{green!10}{14,26}
\Body
\toprule
Method & \multicolumn{2}{c}{BCB} & \multicolumn{2}{c}{MSQ} & \multicolumn{2}{c}{PMem} & \multicolumn{2}{c}{Quality} & \multicolumn{2}{c}{Efficiency} \\
\cmidrule(lr){2-3}\cmidrule(lr){4-5}\cmidrule(lr){6-7}\cmidrule(lr){8-9}\cmidrule(lr){10-11}
& Score $\uparrow$ & TTFT $\downarrow$ & Score $\uparrow$ & TTFT $\downarrow$ & Score $\uparrow$ & TTFT $\downarrow$ & Avg $\uparrow$ & $\Delta$ & Geo. $\downarrow$ & Speedup $\uparrow$ \\
\midrule
\multicolumn{11}{c}{\textbf{Qwen3-4B}} \\
Full & 44.53 & 420.77 & 61.01 & 39.06 & 53.28 & 62.38 & 52.94 & -- & 100.83 & $1.00\times$ \\
PIC & 39.58 & 99.60 & 48.69 & 35.51 & 50.34 & 47.56 & 46.20 & $-6.74$ & 55.20 & $1.83\times$ \\
\multicolumn{11}{c}{\textit{Training-free}} \\
APE & 39.58 & 114.59 & 41.02 & 39.17 & 43.72 & 51.92 & 41.44 & $-11.50$ & 61.54 & $1.64\times$ \\
EPIC & 44.08 & \underline{100.57} & 54.21 & 50.76 & 50.80 & 67.29 & 49.70 & $-3.24$ & 70.03 & $1.44\times$ \\
CacheBlend & \textbf{45.13} & 194.81 & 54.82 & 52.43 & \textbf{53.92} & 69.73 & 51.29 & $-1.65$ & 89.30 & $1.13\times$ \\
AgentKVShift & 42.13 & 196.54 & 52.62 & 68.78 & 47.42 & 90.25 & 47.39 & $-5.55$ & 106.85 & $0.94\times$ \\
\multicolumn{11}{c}{\textit{Training-based}} \\
KVLink & \underline{44.98} & 165.04 & \textbf{59.93} & 98.70 & \underline{52.54} & 109.94 & \textbf{52.48} & $-0.46$ & 121.44 & $0.83\times$ \\
COMB & 35.68 & \textbf{38.29} & 12.70 & \textbf{28.68} & 32.22 & \textbf{43.49} & 26.87 & $-26.07$ & \textbf{36.28} & $\mathbf{2.78\times}$ \\
\textbf{\method} & 44.08 & 101.27 & \underline{58.06} & \underline{36.71} & 51.86 & \underline{47.75} & \underline{51.33} & $-1.61$ & \underline{56.20} & $\underline{1.79\times}$ \\
\midrule
\multicolumn{11}{c}{\textbf{Qwen3-8B}} \\
Full & 50.22 & 660.41 & 63.28 & 62.89 & 43.90 & 99.14 & 52.47 & -- & 160.28 & $1.00\times$ \\
PIC & 47.08 & 113.82 & 50.00 & 35.33 & 35.50 & 68.60 & 44.19 & $-8.27$ & 65.10 & $2.46\times$ \\
\multicolumn{11}{c}{\textit{Training-free}} \\
APE & 46.63 & 123.16 & 44.94 & 39.60 & \underline{44.86} & 73.10 & 45.48 & $-6.99$ & 70.91 & $2.26\times$ \\
EPIC & 48.58 & \underline{110.50} & 56.71 & 50.79 & 42.40 & 88.41 & 49.23 & $-3.24$ & 79.17 & $2.02\times$ \\
CacheBlend & 48.43 & 249.41 & 58.12 & 54.92 & 42.78 & 96.36 & 49.78 & $-2.69$ & 109.69 & $1.46\times$ \\
AgentKVShift & 47.53 & 246.76 & 55.58 & 68.88 & 40.92 & 106.07 & 48.01 & $-4.46$ & 121.71 & $1.32\times$ \\
\multicolumn{11}{c}{\textit{Training-based}} \\
KVLink & \underline{50.22} & 182.91 & \underline{61.20} & 102.70 & \textbf{46.56} & 143.02 & \textbf{52.66} & $+0.19$ & 139.02 & $1.15\times$ \\
COMB & 42.13 & \textbf{48.17} & 14.90 & \textbf{31.02} & 31.62 & \textbf{67.51} & 29.55 & $-22.92$ & \textbf{46.55} & $\mathbf{3.44\times}$ \\
\textbf{\method} & \textbf{50.52} & 113.19 & \textbf{61.22} & \underline{36.33} & 44.14 & \underline{68.10} & \underline{51.96} & $-0.51$ & \underline{65.42} & $\underline{2.45\times}$ \\
\bottomrule
\end{NiceTabular}
\end{adjustbox}
\endgroup
\end{table}

\clearpage

\begin{table}[!htbp]
\centering
\caption{\textbf{Out-of-domain quality and efficiency.}
Scores are accuracy (\%) for TQA and CHP. TTFT is median warm-cache latency in milliseconds. Avg is the arithmetic mean of scores; $\Delta$ is its change from Full in percentage points. Geo. is the geometric mean of TTFT, and speedup is relative to Full. Full and PIC are excluded from ranking; bold and underlined entries mark the best and second-best results among the other methods, including ties.}
\label{tab:app-ood-results}
\begingroup
\small
\setlength{\tabcolsep}{3pt}
\renewcommand{\arraystretch}{1.10}
\begin{adjustbox}{max width=\linewidth}
\begin{NiceTabular}{@{}l*{8}{r}@{}}
\CodeBefore
    \rowcolor{black!6}{3,15}
    \rowcolor{black!6}{6,11,18,23}
    \rowcolor{green!10}{14,26}
\Body
\toprule
Method & \multicolumn{2}{c}{TQA} & \multicolumn{2}{c}{CHP} & \multicolumn{2}{c}{Quality} & \multicolumn{2}{c}{Efficiency} \\
\cmidrule(lr){2-3}\cmidrule(lr){4-5}\cmidrule(lr){6-7}\cmidrule(lr){8-9}
& Score $\uparrow$ & TTFT $\downarrow$ & Score $\uparrow$ & TTFT $\downarrow$ & Avg $\uparrow$ & $\Delta$ & Geo. $\downarrow$ & Speedup $\uparrow$ \\
\midrule
\multicolumn{9}{c}{\textbf{Qwen3-4B}} \\
Full & 71.22 & 89.08 & 69.06 & 139.01 & 70.14 & -- & 111.28 & $1.00\times$ \\
PIC & 66.80 & 38.14 & 64.13 & 46.70 & 65.47 & $-4.68$ & 42.20 & $2.64\times$ \\
\multicolumn{9}{c}{\textit{Training-free}} \\
APE & 59.57 & 42.88 & 60.54 & 49.83 & 60.06 & $-10.09$ & 46.22 & $2.41\times$ \\
EPIC & 67.87 & 46.73 & 66.82 & 55.23 & 67.35 & $-2.80$ & 50.80 & $2.19\times$ \\
CacheBlend & \underline{68.14} & 50.95 & \textbf{68.61} & 75.92 & \underline{68.38} & $-1.77$ & 62.19 & $1.79\times$ \\
AgentKVShift & 67.07 & 61.59 & 65.47 & 79.94 & 66.27 & $-3.87$ & 70.17 & $1.59\times$ \\
\multicolumn{9}{c}{\textit{Training-based}} \\
KVLink & 66.93 & 68.96 & \underline{68.16} & 78.02 & 67.55 & $-2.60$ & 73.35 & $1.52\times$ \\
COMB & 47.52 & \textbf{30.78} & 54.26 & \textbf{31.18} & 50.89 & $-19.25$ & \textbf{30.98} & $\mathbf{3.59\times}$ \\
\textbf{\method} & \textbf{68.81} & \underline{39.80} & \underline{68.16} & 50.62 & \textbf{68.49} & $-1.66$ & \underline{44.89} & $\underline{2.48\times}$ \\
\midrule
\multicolumn{9}{c}{\textbf{Qwen3-8B}} \\
Full & 63.05 & 145.85 & 62.78 & 227.68 & 62.92 & -- & 182.23 & $1.00\times$ \\
PIC & 55.15 & 38.89 & 63.23 & 59.17 & 59.19 & $-3.73$ & 47.97 & $3.80\times$ \\
\multicolumn{9}{c}{\textit{Training-free}} \\
APE & 52.34 & 45.10 & 51.12 & 60.80 & 51.73 & $-11.19$ & 52.36 & $3.48\times$ \\
EPIC & 58.23 & 58.89 & 61.88 & 67.53 & 60.06 & $-2.86$ & 63.06 & $2.89\times$ \\
CacheBlend & 57.56 & 66.42 & 42.60 & 101.32 & 50.08 & $-12.84$ & 82.03 & $2.22\times$ \\
AgentKVShift & \textbf{61.58} & 80.25 & 64.13 & 105.39 & \textbf{62.86} & $-0.06$ & 91.96 & $1.98\times$ \\
\multicolumn{9}{c}{\textit{Training-based}} \\
KVLink & 58.50 & 70.52 & \textbf{66.37} & 92.46 & 62.44 & $-0.48$ & 80.75 & $2.26\times$ \\
COMB & 42.84 & \underline{42.24} & 51.57 & \textbf{44.33} & 47.21 & $-15.71$ & \textbf{43.27} & $\mathbf{4.21\times}$ \\
\textbf{\method} & \underline{60.37} & \textbf{39.80} & \underline{65.02} & \underline{59.98} & \underline{62.70} & $-0.22$ & \underline{48.86} & $\underline{3.73\times}$ \\
\bottomrule
\end{NiceTabular}
\end{adjustbox}
\endgroup
\end{table}

\clearpage
\section{Dataset Examples}
\label{app:dataset-examples}

We show one relevant-only example for each artifact type: skills from BigCodeBench, documents from MuSiQue, and memory messages from PersonaMem-v2. Artifacts are cached independently, while current questions and task instructions remain online. All skill, document, and memory contents are shown in full. Only line wrapping and typographic punctuation are normalized. Block headings, reference answers, reference code, and tests are provided for illustration and are not added to the evaluated prompts.

\begingroup
\lstdefinestyle{datasetprompt}{
    language={},
    basicstyle=\ttfamily\footnotesize,
    backgroundcolor=\color{black!3},
    rulecolor=\color{black!18},
    frame=single,
    framerule=0.35pt,
    framesep=4pt,
    xleftmargin=4.4pt,
    xrightmargin=4.4pt,
    breaklines=true,
    breakatwhitespace=true,
    columns=fullflexible,
    keepspaces=true,
    showstringspaces=false,
    numbers=none,
    aboveskip=5pt,
    belowskip=9pt
}

\subsection{BigCodeBench: Reusable Skills}
\label{app:example-bigcodebench}

Example \texttt{bigcodebench\_00186} supplies two annotated skills: \texttt{geopy} for geographic distances and \texttt{Folium} for map construction. Each complete skill is a separate cached artifact; both are reproduced below without omissions. The two skill blocks precede the task in a single user message, introduced by \texttt{Relevant Skill:} and separated by \texttt{---}.

\Needspace{4\baselineskip}
\noindent\textbf{Cached artifact 1: geopy skill (full text).}
\begin{lstlisting}[style=datasetprompt,emptylines=0]
# geopy -- Distance Calculation and Geocoding

## Geodesic Distance

`geodesic` takes coordinate tuples `(lat, lon)` -- never dicts or lists with named keys:

```python
from geopy.distance import geodesic

dist = geodesic((48.8566, 2.3522), (51.5074, -0.1278)).kilometers
# .miles and .meters are also available
```

## Pairwise Distances Between Locations

When the input is a dict like `{'City': {'Lat': ..., 'Lon': ...}, ...}`, extract tuples before calling `geodesic`. The canonical pattern uses a flat list of `(name, lat, lon)` tuples:

```python
def compute_distances(dic):
    if not dic:
        raise ValueError("Input dictionary is empty.")
    locations = [(k, v['Lat'], v['Lon']) for k, v in dic.items()]
    distances = {}
    for i in range(len(locations)):
        for j in range(i + 1, len(locations)):
            d = geodesic(
                (locations[i][1], locations[i][2]),
                (locations[j][1], locations[j][2])
            ).kilometers
            distances[(locations[i][0], locations[j][0])] = d
    return distances
```

## Geocoding with Photon

`Photon` converts address strings to coordinates. Always pass a `user_agent` string:

```python
from geopy.geocoders import Photon

geolocator = Photon(user_agent="geoapiExercises")
result = geolocator.geocode("Berlin, Germany")
lat, lon = result.latitude, result.longitude
```

## Mixed Input (coordinates or address strings)

When input values are either coordinate dicts `{'Lat': ..., 'Lon': ...}` or plain address strings, use `isinstance` to distinguish them. Instantiate the geolocator **once** outside the loop, and call `geocode()` only for string values -- not for dict values:

```python
def preprocess_locations(dic):
    geolocator = Photon(user_agent="geoapiExercises")
    locations = []
    for name, value in dic.items():
        if isinstance(value, dict) and 'Lat' in value and 'Lon' in value:
            lat, lon = value['Lat'], value['Lon']
        elif isinstance(value, str):
            result = geolocator.geocode(value)
            lat, lon = result.latitude, result.longitude
        else:
            raise ValueError(f"Invalid location format for '{name}': {value}")
        locations.append({'Location': name, 'Lat': lat, 'Lon': lon})
    return locations
```

Calling `geocode()` on a dict value instead of only on string values causes the geocoder to be invoked the wrong number of times.
\end{lstlisting}

\Needspace{4\baselineskip}
\noindent\textbf{Cached artifact 2: Folium skill (full text).}
\begin{lstlisting}[style=datasetprompt,emptylines=0]
# Folium -- Interactive Map Creation

Folium creates Leaflet.js maps in Python. Core pattern: create `folium.Map`, add `folium.Marker` objects, return the map.

## Creating a Map and Adding Markers

`folium.Map(location=...)` and `folium.Marker(location=...)` both require a sequence `[lat, lon]` -- **not** a dict. When data uses `{'Lat': ..., 'Lon': ...}` keys, extract them explicitly:

```python
import folium

m = folium.Map(location=[lat, lon], zoom_start=4)

for name, coords in dic.items():
    folium.Marker(
        location=[coords['Lat'], coords['Lon']],
        popup=name
    ).add_to(m)

return m
```

## Building a DataFrame then Mapping Cities

When coordinates are generated with a seeded RNG, generate lon and lat **per city in a loop** -- one `np.random.uniform` for longitude then one for latitude. Vectorized bulk generation consumes the RNG in a different order:

```python
import pandas as pd
import numpy as np
import folium

def make_city_map(dic, cities):
    lon_min, lon_max = dic['Lon']
    lat_min, lat_max = dic['Lat']

    data = {'City': [], 'Longitude': [], 'Latitude': []}
    for city in cities:
        data['City'].append(city)
        data['Longitude'].append(np.random.uniform(lon_min, lon_max))
        data['Latitude'].append(np.random.uniform(lat_min, lat_max))

    df = pd.DataFrame(data)
    m = folium.Map(location=[df['Latitude'].iloc[0], df['Longitude'].iloc[0]], zoom_start=2)
    for _, row in df.iterrows():
        folium.Marker([row['Latitude'], row['Longitude']], popup=row['City']).add_to(m)
    return m, df
```

## Mixing Dict Coordinates and Address Strings

When input values can be `{'Lat': ..., 'Lon': ...}` dicts or plain address strings, use `isinstance` to distinguish them. Call `geocode()` only for string values -- not for dict values. Instantiate the geolocator once outside the loop:

```python
from geopy.geocoders import Photon

def map_locations(dic):
    geolocator = Photon(user_agent="geoapiExercises")
    preprocessed = []
    for name, value in dic.items():
        if isinstance(value, dict) and 'Lat' in value and 'Lon' in value:
            lat, lon = value['Lat'], value['Lon']
        elif isinstance(value, str):
            result = geolocator.geocode(value)
            lat, lon = result.latitude, result.longitude
        else:
            raise ValueError(f"Invalid location format for '{name}': {value}")
        preprocessed.append({'Location': name, 'Lat': lat, 'Lon': lon})

    locations_df = pd.DataFrame(preprocessed)
    first = locations_df.iloc[0]
    m = folium.Map(location=[first['Lat'], first['Lon']], zoom_start=4)
    for _, row in locations_df.iterrows():
        folium.Marker([row['Lat'], row['Lon']], popup=row['Location']).add_to(m)
    return m
```

## Computing Pairwise Geodesic Distances

Extract `(lat, lon)` tuples before calling `geodesic`. The canonical approach uses a flat list of tuples indexed by position:

```python
from geopy.distance import geodesic

locations = [(name, v['Lat'], v['Lon']) for name, v in dic.items()]
distances = {}
for i in range(len(locations)):
    for j in range(i + 1, len(locations)):
        d = geodesic(
            (locations[i][1], locations[i][2]),
            (locations[j][1], locations[j][2])
        ).kilometers
        distances[(locations[i][0], locations[j][0])] = d
```

`geodesic()` takes `(lat, lon)` tuples -- never a dict.

## Map Object Reference

`m.location` returns the center as `[lat, lon]`. The object type is `folium.folium.Map`.
\end{lstlisting}

\Needspace{4\baselineskip}
\noindent\textbf{Online task (same user message).}
\begin{lstlisting}[style=datasetprompt,emptylines=0]
Generates a Folium map with markers for specified locations and calculates the geodesic distances between each pair of locations.
The function should raise the exception for: ValueError: If the input dictionary is empty.
The function should output with:
    tuple: A tuple containing a Folium map object and a dictionary with pairs of location
    names as keys and their distances in kilometers as values.
You should write self-contained code starting with:
```
from geopy.distance import geodesic
import folium
def task_func(dic):
```
\end{lstlisting}

\Needspace{4\baselineskip}
\noindent\textbf{Evaluation target.}
The generated \texttt{task\_func} implementation is evaluated against the task's unit tests. It must return the map and pairwise-distance dictionary and handle the specified empty-input error. The reference implementation and tests are not included in the prompt.

\Needspace{24\baselineskip}
\noindent\textbf{Reference implementation (evaluation only).}
This is the prepared dataset's reference code, not a generated model
response. It is constructed from the saved code prompt and solution.
\begin{lstlisting}[style=datasetprompt,emptylines=0]
from geopy.distance import geodesic
import folium
def task_func(dic):
    if not dic:
        raise ValueError("Input dictionary is empty.")
    locations = [(k, v['Lat'], v['Lon']) for k, v in dic.items()]
    distances = {}

    folium_map = folium.Map(location=[locations[0][1], locations[0][2]], zoom_start=4)

    for i in range(len(locations)):
        folium.Marker([locations[i][1], locations[i][2]], popup=locations[i][0]).add_to(folium_map)

        for j in range(i + 1, len(locations)):
            distance = geodesic((locations[i][1], locations[i][2]), (locations[j][1], locations[j][2])).kilometers
            distances[(locations[i][0], locations[j][0])] = distance

    return folium_map, distances
\end{lstlisting}

\Needspace{6\baselineskip}
\noindent\textbf{Complete task-specific unit tests (evaluation only).} The following is the full saved test suite. During scoring, the generated function is made available to these tests by the benchmark evaluator; the suite is not part of the model's request.
\begin{lstlisting}[style=datasetprompt,emptylines=0]
import unittest
from unittest.mock import patch
import folium  # Assuming the function task_func and folium are imported or defined appropriately.
class TestCases(unittest.TestCase):
    def test_return_type(self):
        """Test that the function returns a tuple with a map and a dictionary."""
        result = task_func({'Loc1': {'Lat': 0, 'Lon': 0}, 'Loc2': {'Lat': 1, 'Lon': 1}})
        self.assertIsInstance(result, tuple)
        self.assertIsInstance(result[0], folium.folium.Map)
        self.assertIsInstance(result[1], dict)
    def test_distances_calculation(self):
        """Test the accuracy of the distance calculation. Assumes the distance is reasonable for nearby points."""
        _, distances = task_func({'Loc1': {'Lat': 0, 'Lon': 0}, 'Loc2': {'Lat': 0, 'Lon': 1}})
        self.assertTrue(0 < distances[('Loc1', 'Loc2')] < 200)  # Rough check for distance in kilometers
    def test_multiple_locations(self):
        """Test functionality with multiple locations."""
        _, distances = task_func({'Loc1': {'Lat': 0, 'Lon': 0}, 'Loc2': {'Lat': 0, 'Lon': 1}, 'Loc3': {'Lat': 1, 'Lon': 1}})
        self.assertEqual(len(distances), 3)  # Expecting 3 pairs of locations
    def test_marker_addition(self):
        """Test that markers are correctly added to the map. Assumes 1 TileLayer present."""
        folium_map, _ = task_func({'Loc1': {'Lat': 0, 'Lon': 0}})
        self.assertEqual(len(folium_map._children), 2)  # One for TileLayer and one for Marker
\end{lstlisting}

\Needspace{18\baselineskip}
\noindent\textbf{Unit tests (continued).}
\begin{lstlisting}[style=datasetprompt,emptylines=0]
    @patch('geopy.distance.geodesic')
    def test_distance_dict_structure(self, mock_geodesic):
        """Ensure the distance dictionary has the correct key-value structure."""
        mock_geodesic.return_value.kilometers = 100  # Mock distance as 100 km
        _, distances = task_func({'Loc1': {'Lat': 0, 'Lon': 0}, 'Loc2': {'Lat': 0, 'Lon': 1}})
        self.assertTrue(all(isinstance(key, tuple) and isinstance(value, float) for key, value in distances.items()))
    def test_empty_input(self):
        """Test function behavior with an empty dictionary input raises ValueError."""
        with self.assertRaises(ValueError):
            task_func({})
    def test_single_location(self):
        """Test handling of a single location input."""
        folium_map, distances = task_func({'Loc1': {'Lat': 0, 'Lon': 0}})
        self.assertEqual(len(distances), 0)  # No distances calculated
        self.assertEqual(len(folium_map._children), 2)  # One for TileLayer and one for Marker
    def test_negative_lat_lon(self):
        """Test handling of negative latitude and longitude values."""
        _, distances = task_func({'Loc1': {'Lat': -34, 'Lon': -58}, 'Loc2': {'Lat': -33, 'Lon': -70}})
        self.assertTrue(all(value >= 0 for value in distances.values()))  # Distance should be positive
    def test_large_distance_calculation(self):
        """Test accuracy for large distances, e.g., antipodal points."""
        _, distances = task_func({'Loc1': {'Lat': 0, 'Lon': 0}, 'Loc2': {'Lat': 0, 'Lon': 180}})
        self.assertTrue(distances[('Loc1', 'Loc2')] > 10000)  # Expecting a large distance
\end{lstlisting}

\noindent\textbf{What the example illustrates.}
The task combines two reusable library skills: coordinate tuples feed geodesic-distance calculations, while Folium creates the map and markers. Correctness is determined by the task-specific tests, rather than by textual agreement with the reference implementation.

\clearpage
\subsection{MuSiQue: Supporting Documents}
\label{app:example-musique}

Example \texttt{2hop\_\_460946\_294723} contains two annotated supporting documents. Each cached artifact includes its document index, title, and paragraph. The original indices and relative order are retained; the other candidate documents are excluded. The system instruction, \texttt{Documents:} heading, question, and answer-format instruction are outside the cached document spans.

\Needspace{4\baselineskip}
\noindent\textbf{System prompt.}
\begin{lstlisting}[style=datasetprompt]
You answer multi-hop questions using only the provided documents.
\end{lstlisting}

\Needspace{4\baselineskip}
\noindent\textbf{Cached artifact 1: supporting document.}
\begin{lstlisting}[style=datasetprompt]
[Document 5]
Title: Miquette Giraudy
Miquette Giraudy (born 9 February 1953, Nice, France) is a keyboard player and vocalist, best known for her work in Gong and with her partner Steve Hillage. She and Hillage currently form the core of the ambient band System 7. In addition to her performances in music, she has also worked as an actress, film editor and writer. In each role, she has used different stage names.
\end{lstlisting}

\Needspace{4\baselineskip}
\noindent\textbf{Cached artifact 2: supporting document.}
\begin{lstlisting}[style=datasetprompt]
[Document 10]
Title: Green (Steve Hillage album)
Green is the fourth studio album by British progressive rock musician Steve Hillage. Written in spring 1977 at the same time as his previous album, the funk-inflected "Motivation Radio" (1977), "Green" was originally going to be released as "The Green Album" as a companion to "The Red Album" (the originally intended name for "Motivation Radio"). However, this plan was dropped and after a US tour in late 1977, "Green" was recorded alone, primarily in Dorking, Surrey, and in London.
\end{lstlisting}

\Needspace{4\baselineskip}
\noindent\textbf{Online question and answer-format instruction.}
\begin{lstlisting}[style=datasetprompt]
Question: Who is the spouse of the Green performer?

Give a concise answer. Put the final answer on its own last line exactly as `ANSWER: <answer>`.
\end{lstlisting}

\Needspace{4\baselineskip}
\noindent\textbf{Reference answer (shown in the requested output format).}
\begin{lstlisting}[style=datasetprompt]
ANSWER: Miquette Giraudy
\end{lstlisting}

\noindent\textbf{Evidence linking.}
The Green document identifies Steve Hillage as the performer; the other document names Miquette Giraudy as his partner. The dataset reference answer therefore requires linking information across the two documents.

\clearpage
\subsection{PersonaMem-v2: Relevant Memory Messages}
\label{app:example-personamem}

Example \texttt{personamem-v2:689:04277} supplies two messages from the annotated related conversation snippet. Each complete historical message, including its role, forms a separate artifact. We preserve their order and the saved answer-option mapping; unannotated history and persona background are not added. The current question and subsequent system message remain online.

\Needspace{4\baselineskip}
\noindent\textbf{Cached artifact 1: historical user message.}
\begin{lstlisting}[style=datasetprompt]
Can you help me make this sound a bit smoother?
\end{lstlisting}

\Needspace{4\baselineskip}
\noindent\textbf{Cached artifact 2: historical assistant message.}
\begin{lstlisting}[style=datasetprompt]
It's been raining all afternoon, so I spent the day indoors working on new drum fills and playing a few 2K matches. It's been a nice, laid-back day.
\end{lstlisting}

\Needspace{4\baselineskip}
\noindent\textbf{Online user question.}
\begin{lstlisting}[style=datasetprompt]
What are some cozy and relaxing activities I could do at home when I have a free afternoon? Please recall my related preferences from our conversation history to give personalized responses.
\end{lstlisting}

\Needspace{4\baselineskip}
\noindent\textbf{Online system message: options and answer-format instruction.}
\begin{lstlisting}[style=datasetprompt]
Please choose the best answer from the following options:

A. On a free afternoon, you could make your favorite warm drink, put on some calming music, and work on a personal hobby. Later, you could read a good book or watch a movie to relax before dinner.
B. On a quiet afternoon, you could brew a fresh cup of masala chai, listen to some Bollywood classics, and help your mom with a home-cooked dinner. Later, you could chat with family about cricket matches and end the day watching a classic Shah Rukh Khan movie.
C. On a rainy afternoon, you could make some chai, put on your favorite jazz playlist, and sketch some new sneaker designs while listening to the rain. Later, you could practice a few new drum fills and then wind down by watching NBA highlights or an inspiring drumming tutorial.
D. On a free afternoon, you could make some iced lemonade, play around with a new pasta recipe, and then spend time reading travel blogs. Later, you could try learning a bit of Italian on a language app and finish the day with a feel-good movie.

Think step by step about which answer best fits the user's query and conversation context. Provide your reasoning first, then give your final answer as 'Final Answer: [Letter]'
\end{lstlisting}

\Needspace{4\baselineskip}
\noindent\textbf{Reference option (shown in the requested output format).}
\begin{lstlisting}[style=datasetprompt]
Final Answer: C
\end{lstlisting}

\noindent\textbf{Memory-dependent choice.}
The historical assistant message mentions drum practice and basketball video games. The annotated answer is option C, which includes drum practice; the prepared option order and label are retained exactly. This is a multiple-choice reference, not a model-generated response.

\endgroup

\end{document}